\documentclass[10pt,twocolumn,letterpaper]{article}
\pdfoutput=1 
\usepackage[utf8]{inputenc}
\usepackage{cvpr}
\usepackage{times}
\usepackage{epsfig}
\usepackage{graphicx}
\usepackage{amsmath}
\usepackage{amssymb}
\usepackage{caption}
\usepackage{mdframed}
\usepackage{bm}
\usepackage{graphicx}
\usepackage{float}
\usepackage{caption}
\usepackage{subcaption}
\usepackage{authblk}
\def\alert#1{\textcolor{red}{#1}}

\newmdenv[innerlinewidth=0.8pt, roundcorner=0.5pt,linecolor=mycolor,innerleftmargin=4pt,
innerrightmargin=4pt,innertopmargin=4pt,innerbottommargin=4pt]{mybox}
\definecolor{mycolor}{rgb}{1, 0, 0}

\DeclareMathOperator*{\argmin}{arg\,min}

\def\one{{\bm 1}}
\def\zero{{\bm 0}}

\usepackage[pagebackref=true,breaklinks=true,letterpaper=true,colorlinks,bookmarks=false]{hyperref}

\cvprfinalcopy 

\begin{document}
\title{Monocular Depth Parameterizing Networks}

\author[1]{Patrik Persson}
\author[1]{Linn Öström}
\author[1,2]{Carl Olsson}
\affil[1]{Lund University, Sweden}
\affil[2]{Chalmers University of Technology, Sweden}

\maketitle

\begin{abstract}
Monocular depth estimation is a highly challenging problem that is often addressed with deep neural networks. While these are able to use recognition of image features to predict reasonably looking depth maps the result often has low metric accuracy. In contrast traditional stereo methods using multiple cameras provide highly accurate estimation when pixel matching is possible.

In this work we propose to combine the two approaches leveraging their respective strengths. For this purpose we propose a network structure that given an image provides a parameterization of a set of depth maps with feasible shapes. Optimizing over the parameterization then allows us to search the shapes for a photo consistent solution with respect to other images. 
This allows us to enforce geometric properties that are difficult to observe in single image as well as relaxes the learning problem allowing us to use relatively small networks.
Our experimental evaluation shows that our method generates more accurate depth maps and generalizes better than competing state-of-the-art approaches. 
\end{abstract}

\section{Introduction}\label{sec:intro}

Dense depth or disparity estimation is a classical problem in computer vision \cite{birchfield-tomasi-cvpr-1999,middlebury2,veksler-cvpr-2005,bleyer2010surface}.
Traditional methods use stereo (or multi-camera) setups and attempt to match every pixel in the reference image to a corresponding pixel in a neighbouring image using appearance cues. While the accuracy of the recovered depth is often very high for correctly matched pixels ambiguous texture can degrade the matching and often leads to a noisy depth map.

\begin{figure}[ht!]
    \centering
  \resizebox{0.4\textwidth}{!}{   \includegraphics[width = 18mm]{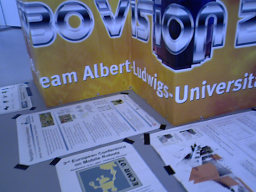}    
 \includegraphics[width = 22mm, trim = 20 180 80 80]{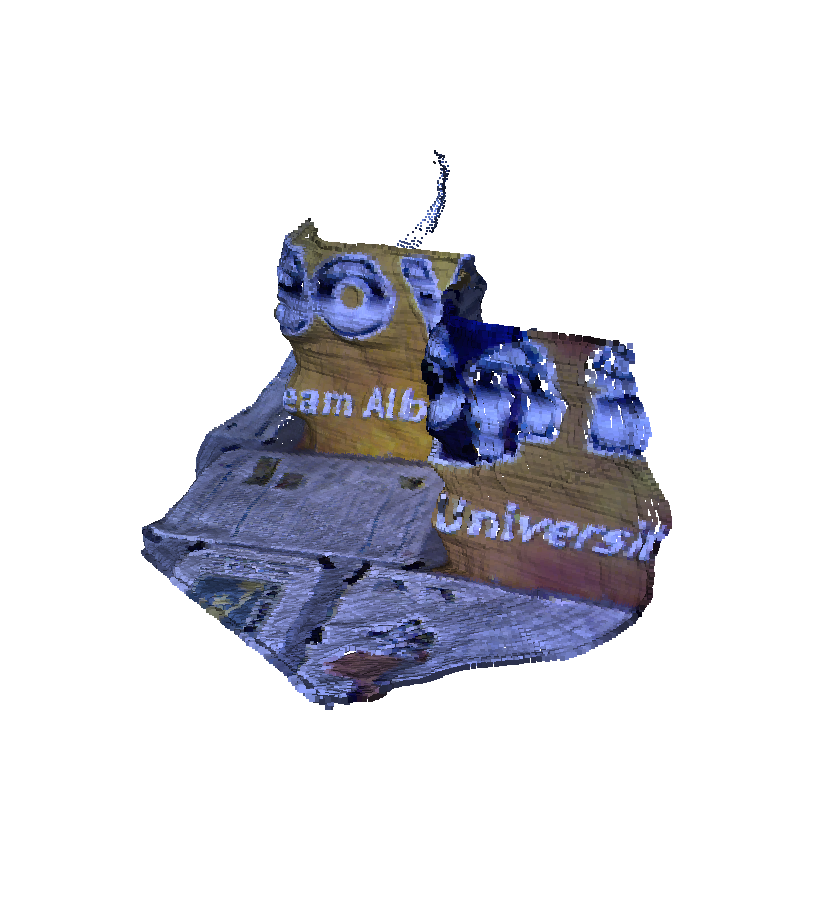}
 }
 \resizebox{0.4\textwidth}{!}{\vspace{-6cm}\hspace{3cm}\footnotesize{\textbf{Image}}\hspace{3cm}\footnotesize{\textbf{Ours}}\hspace{1cm}}
 \resizebox{0.4\textwidth}{!}{   \includegraphics[width = 18mm, trim = 100 80 80 140]{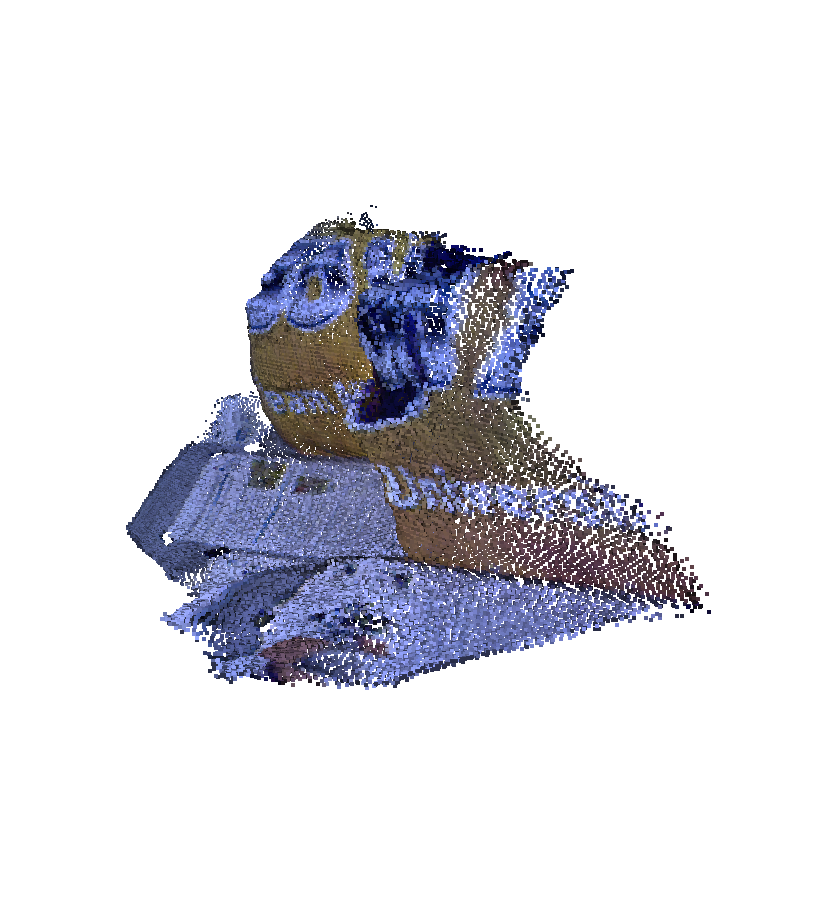}\includegraphics[width = 18mm, trim = 60 100 100 100]{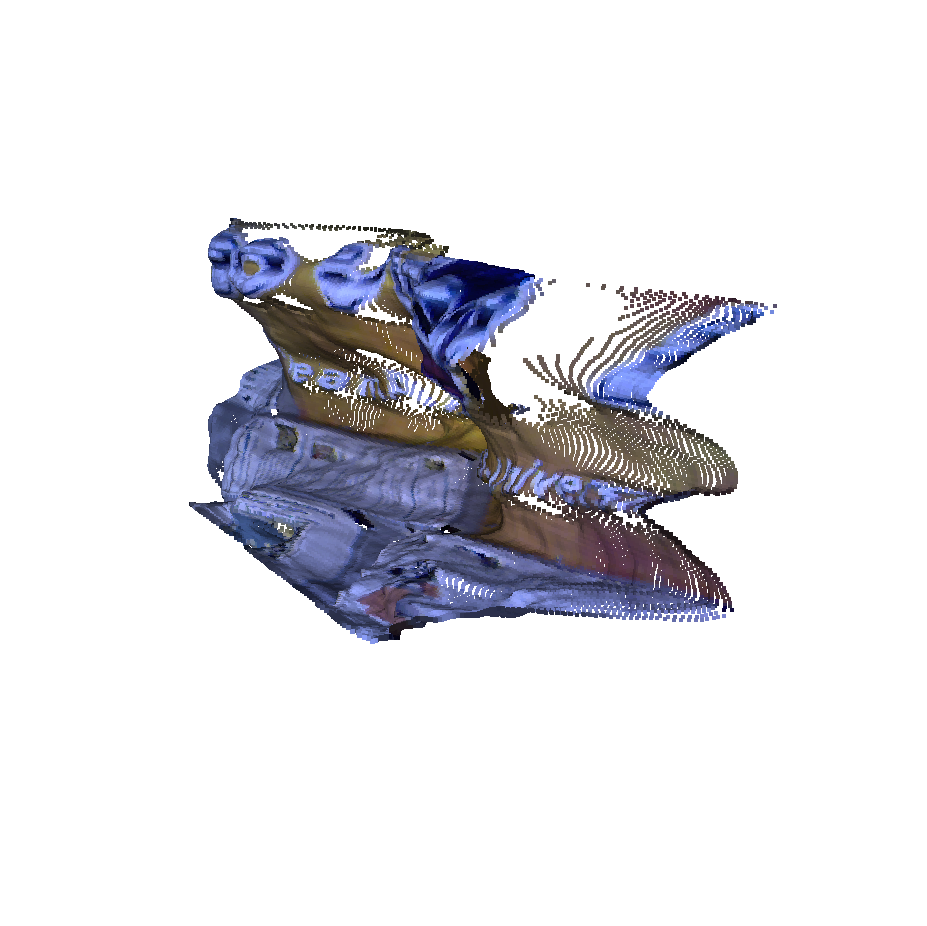}
}
\\
\vspace{-0.8cm}
 \resizebox{0.45\textwidth}{!}{\vspace{-6cm}\hspace{4cm}\footnotesize{\textbf{DeepFactors \cite{Czarnowski:2020:10.1109/lra.2020.2965415}}}\hspace{3cm}\footnotesize{\textbf{MegaDepth \cite{li2018megadepth}}}}
    \caption{Our monocular depth parameterizing network produces high quality and detailed reconstructions. We achieve realistic-looking depth maps and reconstructions for a large variety of scenes.  }
    \label{fig:bad_momodepth}
\end{figure}
To stabilize the result a popular approach is to add geometric regularization terms such as derivative \cite{boykov-etal-pami-2001, kolmogorov-zabih-eccv-2002, felzenswalb-huttenlocher-ijcv-2006} or curvature \cite{liZucker2010,woodford2009,olsson-etal-cvpr-2013} penalties. These can be realized as low order potentials in a conditional random field and efficient inference can be performed with move-making \cite{boykov-etal-pami-2001,Lempitsky-etal-pami-2010,veksler-cvpr-2007} or message passing algorithms \cite{kolmogorov06,felzenswalb-huttenlocher-ijcv-2006}. While this kind of prior can drastically improve the 
estimation in ambiguous image regions, they lack any ability to recognize complex geometries and are basically limited to encouraging piecewise planar or smooth surfaces.

A more recent approach is to use neural networks, trained on examples, to directly infer depth or dense matching \cite{Mayer-etal-cvpr-2016,zhang2019domaininvariant,watson-etal-eccv-2020,chang2018pyramid,Tonioni_2019, Zhou_2017_ICCV,chang2018pyramid, Kendall_2017, Zhang_2019}. 
These models can potentially work better in low textured regions since they  
can learn high level recognition of geometric structures on a larger scale than what traditional methods work on.
On the other hand they can generalize poorly and require lots of training data \cite{watson-etal-eccv-2020,zhang2019domaininvariant,Zhang2019GANet}.
An extreme case of this is monocular depth estimation where a neural network is used to estimate depth from a single image \cite{MegaDepthLi18,eigen2014depth,single_image_in_the_wild,depth_resnet,eigen2014depth}. These networks typically require a huge amount of training data and may generalize poorly \cite{garg2016unsupervised}.
In addition, while these achieve meaningful results with plausible object shapes, the resulting depth maps are often inaccurate because of the ambiguous nature of the problem \cite{nishimura-etal-eccv-2020,huynh-etal-eccv-2020}, see Figure \ref{fig:bad_momodepth}.

In this context it is interesting to think about what variations in the depth map can be explained by the image features.
As a simplified example consider an orthographic camera viewing a known object. It is well known that the absolute distance between the object and the camera cannot be determined from image data \cite{hartley-zisserman-book-2004}. In a sense there is a loss of information due to the projection from 3D to 2D not being injective (one-to-one). Therefore we cannot hope to invert it without supplying additional information such as an extra view.
Hence a network that uses ground truth data where the absolute distance varies will not be able to predict such variations regardless of the amount of training data it is being fed. There may be other factors than the image formation model that limits what information can be extracted from a single image. In practice the size and architecture of the network determines what features can be used for prediction. We refer to information about the depth map that cannot be extracted from the features as the knowledge-gap. 

In this paper we aim to resolve ambiguities of the monocular depth map estimation by combining learning models with traditional geometric formulations. Our goal is to use a neural network to extract a low dimensional shape parameterization from a single image that is flexible enough to allow depth fitting using traditional stereo ques such as photo consistency. To achieve this we add an additional input, which models the knowledge-gap, to a U-net architecture designed to complement the model with the information that is not directly observed in the image. The result is a network that takes as input an image 
and gives a class of possible depth-maps parametrized by the hidden variables. 
We emphasize that in contrast to current monocular depth approaches \cite{MegaDepthLi18,eigen2014depth,single_image_in_the_wild,depth_resnet} we are not aiming for a network that predicts absolute pixel depths but rather one that extracts reliable shape information that reduces the dimensionality of a traditional stereo formulation.
This allows us to use smaller networks, lessening the need for training data.

Our network can be trained without requiring any knowledge of the shape variables, using either images with accompanying depth maps or self-supervised through geometric and photo-metric consistency losses. We show in our experimental evaluation that our approach gives more accurate solutions than what is normally achieved with learning based depth prediction methods.

In summary our main contributions are:
\begin{itemize}
	\item We present a new network for extracting low dimensional shape priors from monocular images.
	\item We show that explicitly modelling the knowledge gap enables training of a flexible model that can be combined with traditional stereo methods.
	\item We show that the model can be trained in a self-supervised end-to-end fashion.
	\item Our empirical evaluation demonstrates that our approach achieves much more accurate depth estimates than stat-of-the-art learning based depth prediction.
\end{itemize}

\subsection{Related Articles}
Until recently, the best practice has been to train monocular depth prediction networks in a supervised manner \cite{eigen2014depth, Czarnowski:2020:10.1109/lra.2020.2965415, code_slam}.
However, these require large amounts of training data and for
this reason the use of multiple view geometry to aid weakly or self-supervised training, has become popular
\cite{garg2016unsupervised,MegaDepthLi18,monodepth17, monodepth2}.
For example, \cite{MegaDepthLi18} produces high quality depth maps from Internet Photos using SfM and multi-view stereo (MVS) methods to achieve strong generalization. A loss function measuring photo-consistency of the source image warped to a neighboring image according to its predicting depth map and the camera geometry is used in \cite{garg2016unsupervised}.
Similarly \cite{monodepth17} uses epipolar geometry to define a suitable loss function.
In \cite{tiwari2020pseudo} the training of a monocular depth network is coupled with a SLAM system to form a self-improving loop. While these methods use geometry to simplify/improve training, the end goal is still a monocular network able to predict depth from the image alone. Such approaches do not help in disambiguating the problem when there is a loss of information due to projection.

To be able to disambiguate 3D structure from a single image, a so called depth attention volume is used in \cite{huynh-etal-eccv-2020} to modify the bottleneck features of an encoder-decoder type architecture to favour planar regions. Another approach is presented in \cite{nishimura-etal-eccv-2020}, where a histogram of pixel depths is supplied to the network in addition to the image. 

The work that is perhaps most similar to our is
\cite{Czarnowski:2020:10.1109/lra.2020.2965415, code_slam} where an image dependent latent scene representation is leaned in a supervised manor, using a depth-map auto-encoder coupled with a U-Net for convolutional image feature extraction. At testing, the depth map encoder is removed resulting in a network that takes a code and outputs a depth map conditioned on the image. They observe that this allows the code to represent local geometry that cannot be directly observed in the image. A search over both code and camera pose is performed during testing. The presented network is however complex and it is trained with ground truth RGB-D images.
The depth decoder is linear suggesting that the prediction might be highly dependent on image features (see Figure 8 of \cite{code_slam}). Our latent variable model is not trained to recreate a depth map through an autoencoder but to complement image features with information needed to predict the depth map. We will show that in our approach we can achieve better depth accuracy using a much simpler network, training setup and using only self-supervision, see Section~\ref{sec:comparison_to_others}.

\section{Parameterizing Feasible Shapes}

While it may be hard for a network to learn exact monocular depth prediction we argue that it is of interest to be able to extract what the network is able to learn, even if this is not sufficient to pinpoint a particular depth map with certainty from a single image. Instead we design our network so that it can be complemented with additional information to resolve ambiguities.

For this purpose we use a network reminiscent of a U-net \cite{Ronneberger_2015}, with a contracting and an expanding part connected with skip connections for accurate localization, but add an additional input that provides the expansion part with additional information that allows it disambiguate. This essentially results in a low dimensional shape model which, at testing time, we search for a photo-consistent solution.
Before we present our network we illustrate the basic principle using a simplified example.

\subsection{Overall Principle}\label{sec:matrix_model}

We consider a matrix formulation with a contractive part, represented by a matrix $F$, which takes an input in the form a row vector $x$ and outputs a feature vector $x F$. 
Here $xF$ represents features extracted from the image. 
To this we concatenate a vector $z$ of latent variables that complements the feature information into a larger feature vector
$\begin{bmatrix}
xF & z
\end{bmatrix}$ and feed the result to an expanding map $G$ giving the output
\begin{equation}
\bar{y}(x,z) = \begin{bmatrix}
xF & z
\end{bmatrix}G = xFG_x + zG_z.
\label{eq:linmodel}
\end{equation}
Here $G_x$ and $G_z$ are the rows of $G$ corresponding to the $x$ and $z$ variables respectively.
Given ground truth data $x_i$ and $y_i$ a feasible loss function would be
\begin{equation}
\sum_i \|x_i F G_x + z_i  G_z- y_i\|^2 = \| X FG_x + ZG_z - Y\|_F^2, 
\label{eq:loss}
\end{equation}
where $X$, $Y$ and $Z$ contain the row stacked vectors $x_i$, $y_i$ and $z_i$ respectively. 
Note that the network is allowed to freely determine $Z$ during training. 

First consider the case when $Z=0$ and suppose that $Y$ cannot be fully recreated using the term $XFG_x$. Since $G_x$ can be selected freely during training this occurs when the feature matrix $XF$ has a lower rank than $Y$, which is either due to an information loss between $Y$ and $X$ or $F$ being to small to extract all relevant information. For simplicity we restrict ourselves to the former situation, the latter is similar. When dividing the matrix $Y$ into two components, $\mathcal{P}Y$ which is the projection of the columns of $Y$
onto the column space of $X$ and $\mathcal{P}_\perp Y = Y-\mathcal{P}Y$,
the training objective reduces to
\begin{equation}
\|XFG_x - \mathcal{P}Y\|^2 + \|\mathcal{P}_\perp Y\|^2.
\end{equation}
The first term can be made to vanish by any choice of $F$ and $G$ such that $FG_x=X^\dagger \mathcal{P}Y$,
where $X^\dagger$ is the pseudo-inverse of $X$. 
The second term is not affected by the choice of $F$ and $G_x$ and represents the variations in the $y_i$ variables that cannot be explained by the inputs $x_i$.

For comparison consider what happens when we allow usage of $Z$. 
If the columns of $Z$ are chosen to be perpendicular to the column space of $X$ the objective \eqref{eq:loss} becomes
\begin{equation}
\|XFG_x - \mathcal{P}Y\|^2 + \|Z G_z - \mathcal{P}_\perp Y\|^2.
\label{eq:Zloss}
\end{equation}
Hence after minimization $Z$ will contain the information missing in the feature space needed to explain the ground truth variations and $G_z$ will have learned how to decode this information.

We give more explanations and details on this matrix model and study its properties in the orthographic projection example in the appendix.

\subsection{Network Structure}
\begin{figure}[htb]
\begin{center}
	\includegraphics[width=75mm]{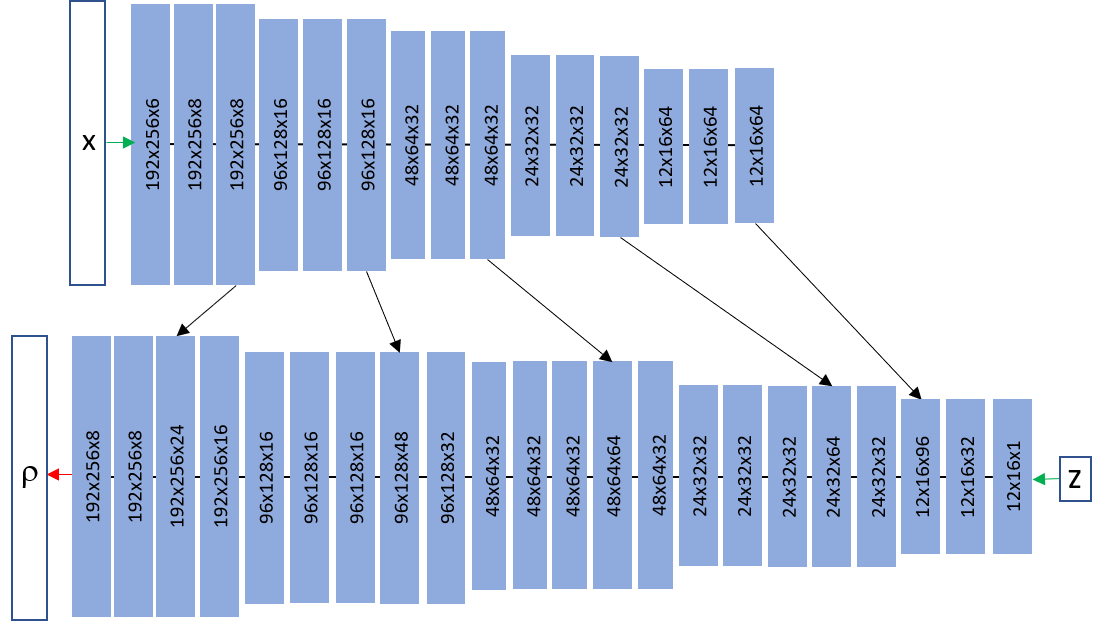}
\end{center}
	\caption{\textbf{Our network architecture}. Both the contracting path (top) and the expanding part (bottom) consists of convolutional layers connected by LeakyReLU. Arrows show skip connections.}
	\label{fig:basic_network}
\end{figure}

The network we propose takes an image $I$, a set of normalized pixel coordinates $Q$ and a vector $z$ and outputs a transformed depth map $\rho$. The basic construction, which is shown in Figure~\ref{fig:basic_network}
consists of a U-net type architecture with the addition of the extra inputs $z$. The contracting component computes features at different resolutions using convolutional layers. They are supplied to the expanding part at corresponding levels. 
The additional input $z$ takes a vector of $192$ values (corresponding to a grid of size $12 \times 16$)
and complements the features with information needed for prediction. 
The input image $x$ and the output map $\rho$ are both of size $192 \times 256$. Hence given an input image our network provides a map from a $192$-dimensional parameter space into a $49152$-dimensional space of depth maps. Compared to a traditional stereo matching formulation 
optimizing over all pixels we thus get a significant reduction of the search space. 
We also remark that due to its simple structure, the number of parameters in our network is relatively small, all in all roughly 400000.

The network can be thought of as a functional $\rho = N(I,Q,z|W)$ which takes an image $I$, normalized pixel coordinates $Q$, the latent variable $z$ and outputs a transformed depth map $\rho$ with values in $[0,1]$. We include the normalized pixel coordinates as input to allow the network to account for different camera calibrations. Similar to \cite{code_slam}, the  output $\rho(p)$ at a pixel $p$ from the network relates to depth $D(p)$ as

\begin{align}
	\rho(p) = \frac{\alpha}{D(p) + \alpha}
\end{align}

where $\alpha$ is the mean depth. This will map the network output from $[0,1]$ to $[0, \infty]$.  The depth map $D(p)$ can then be recovered using

\begin{equation}
D(p) = \frac{\alpha (1 - \rho(p))}{ \rho(p)}.
\end{equation}

By $d$ we denote the functional that takes as input $\rho=N(I,Z,z|W)$ and $\alpha$ and gives the depth map $D$ as output, that is,
$D = d(\rho,\alpha)$.
In this paper we train the network using two different approaches. In the first we use image and ground-truth depth pairs to do supervised learning, see Section~\ref{sec:supervisedtraing}. In the second we assume that no ground-truth depth is available and instead train using only photometric and geometric errors between co-visible images, see Section~\ref{sec:selfsupervisedtraining}. The latter results in self-supervised learning where we learn depth indirectly from the images. This has the benefit that any partially overlapping collection of images can be used to train the network. The training therefore becomes more scalable since no hard to obtain depth ground truth is needed. In addition it makes it easy to calibrate for a new camera since we only need images from this camera to update the network.

\subsection{Testing and Optimization} \label{sec:shape_optimization}
As previously discussed the output of our network can be seen as a family of feasible depth maps parameterized by the latent variables $z$.
In the testing phase we therefore need to perform an optimization over $z$ to find the depth map with the best photometric fit.
For this purpose we define a set of relative transformations $T_{ij}$ that transforms 3D points in camera coordinate system $i$ to $j$.
Given an image $I_i$ and associated depth map $D_i$ and normalized image coordinates $Q_i$ a pixel $p$ in image $i$ is the projection of the 3D point
\begin{equation}
    X_i(p) = \begin{bmatrix}
    Q_i(p) \\ 1
    \end{bmatrix}D_i(p).
\end{equation}
This point is transformed to pixel
$
    q_{ij}(p) = \pi \left(T_{ij} X(p) \right),
$
in image $j$, where $\pi$ is the projection mapping (division with the 3rd coordinate).
For a photoconsistent solution we therefore want
\begin{equation}
\mathcal{L}_{\text{photo}_i} = \lambda_p \sum_{j\neq i} \|M_{ij} \odot (I_j \circ q_{ij} - I_i)\|_\delta,
\label{eq:lphoto}
\end{equation}
to be small. Here $M_{ij}$ is a masking matrix that removes pixels that are not visible (we discuss the generation of this mask in detail below), and $\|\cdot\|_\delta$ is the Huber norm calculated pixel-wise and $\lambda_p = 100$ is a scalar weight factor. Since we are dealing with discrete images the composition $I_j \circ q_{ij}$ involves re-sampling image $j$ at the points $q_{ij}(p)$ using bilinear interpolation. Note also that the above loss depends on $z_i$ through $q_{ij}$. 
We also want the depth maps from different cameras to give consistent 3D point clouds.
The depth $D_{ij}(p)$ of the point $X_i(p)$ in camera $j$ is given by the third coordinate of $T_{ij}X_i(p)$.
We therefore add the loss 
\begin{equation}
\mathcal{L}_{\text{depth}_i} = \lambda_d \sum_{j \neq i} \|\frac{1}{\alpha_j} M_{ij} \odot (D_j \circ q_{ij} - D_{ij})\|_\delta,
\label{eq:ldepth}
\end{equation}
to encourage consistent depth maps. Here $\lambda_d = 10$ is a weight factor and $\alpha_j$ is the mean depth of $D_j$. Division with $\alpha_j$ makes the above term scale invariant. 

We conclude this section by discussing the masking operators $M_{ij}$. 
When transforming one view into another there are inevitably going to be some pixels that cannot be transferred due to limited field of view, chirality and occlusion. Pixels that end up outside the image bounds or have negative depth are easily removed based on their location in the camera coordinate system. However to determine occlusion we also need to consider the current estimate of the depth map.  
For this purpose we compare the predicted depth $D_j$ with the projected depth $D_{ij}$. If there is no occlusion we assume that the error $\Delta_{ij}(p) = D_j\circ q_{ij}(p)-D_{ij}(p)$ is approximately Gaussian distributed (i.i.d over the pixels). We use median absolute deviation to get a robust estimate of the mean and standard deviation of the depth error distribution. We classify pixel $p$ as occluded if the corresponding depth error $\Delta_{ij}(p)$ satisfies
\begin{align}
	\Delta_{ij}(p) \geq \textbf{Median}(\Delta_{ij}) - \tau \cdot \textbf{MAD}(\Delta_{ij}),
\end{align}
where $\tau$ is $4.44$ and the median and median absolute deviation is taken over all pixels in the image.

In addition to invalid and occluded points, we also mask out points with a shallow viewing angle. These occur on surfaces that are very slanted compared to the viewing direction and on the boundary of depth discontinuities. In both situations the pixel and depth values at these points may not be informative or may even be detrimental to the solution. They are therefor masked out. These are found by estimating the normal of the surface point and calculating the angle between normal and viewing direction. If the angle is above a threshold of $85^{\circ}$, the point is masked out.

For optimizing the above model we use an AdaMax  optimizer \cite{adamax} to minimize the loss. 
This is iterated until the relative difference between the current and previous loss falls below a threshold. \footnote{Code is available at \href{https://github.com/patrikperssonmath/MDPN}{https://github.com/patrikperssonmath/MDPN}.}

\begin{figure*}[ht!]
\hspace{-0.3cm}
    \resizebox{0.265\textwidth}{!}{\begin{subfigure}[b]{0.3\textwidth}
    \centering
    \hspace{0.2cm}
 \includegraphics[width = 2cm, trim = 100 100 100 100]{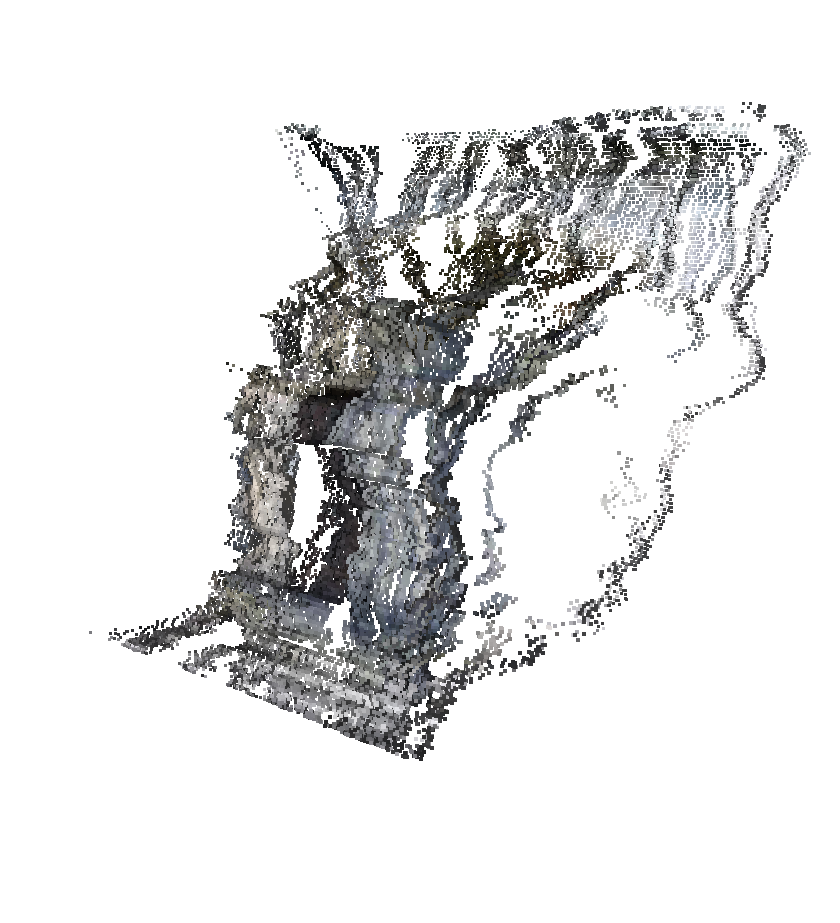}\hspace{0.3cm}
    \includegraphics[width = 2.3cm, trim = 100 100 100 100]{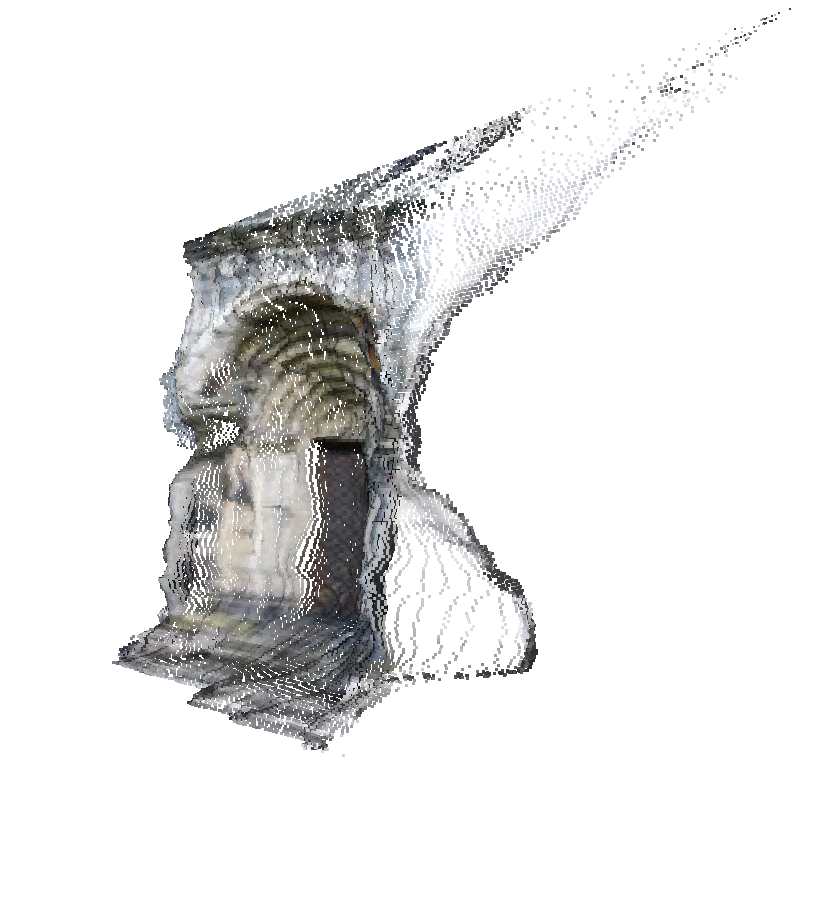}\hspace{-0.2cm}
    \\
     \includegraphics[width = 1.5cm]{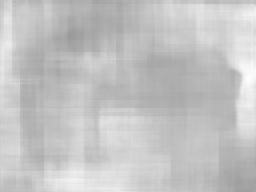}
    \includegraphics[width = 1.5cm]{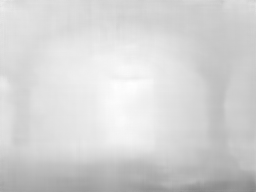}
    \includegraphics[width = 1.5cm]{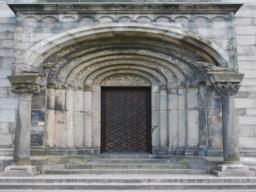}
    \\
    \hspace{0.3cm}
    \footnotesize{\textbf{Mono}} \hspace{0.4cm} \footnotesize{\textbf{Photometric}} \hspace{0.3cm} \footnotesize{\textbf{Input}}\hspace{0.4cm}
    \label{fig:my_label}
    \caption{\textbf{Dataset Door \cite{enqvist-etal-omnivs-2011}.}}
     \end{subfigure}}\hspace{-0.4cm}
\resizebox{0.265\textwidth}{!}{\begin{subfigure}[b]{0.3\textwidth}
    \centering
    \hspace{0.5cm}
 \includegraphics[width = 1.3cm, trim = 120 120 120 120]{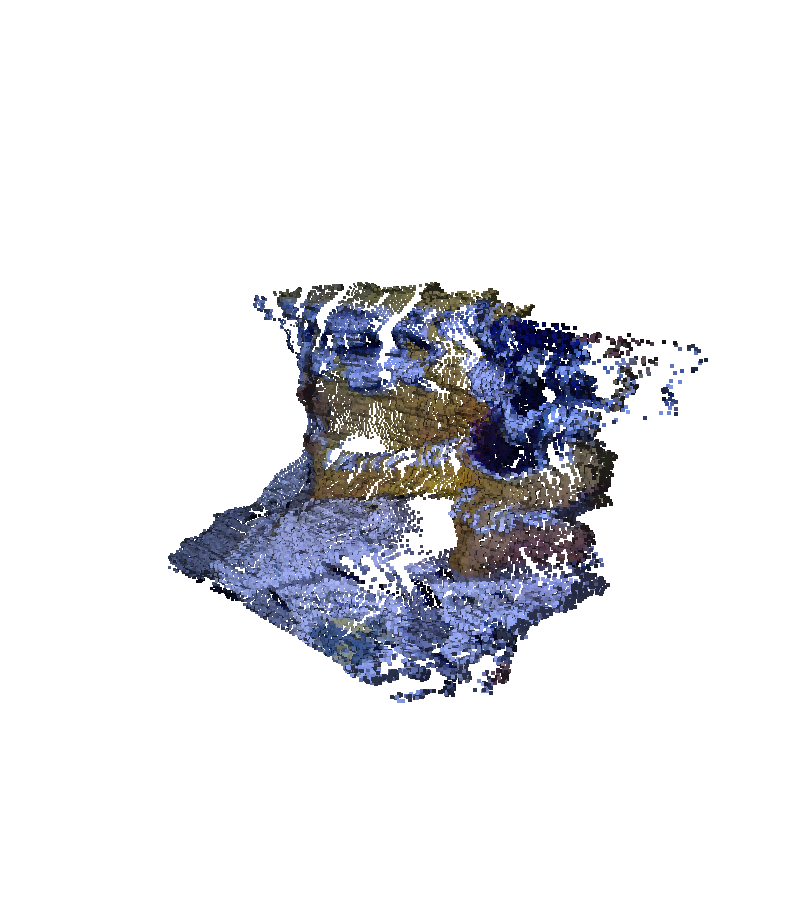}\hspace{0.5cm}
     \includegraphics[width = 1.3cm, trim = 150 150 150 150]{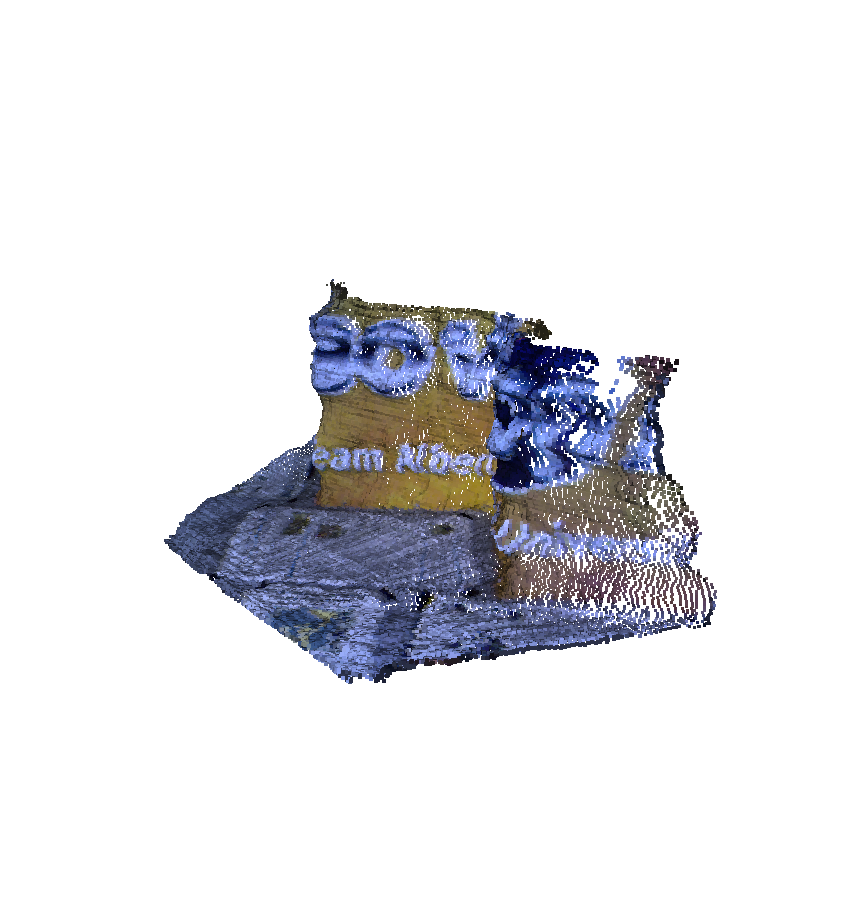}\vspace{0.7cm}
    \\
    \vspace{-1cm}
     \hspace{-0.5cm}
    \includegraphics[width = 1.3cm, trim = 130 130 130 130]{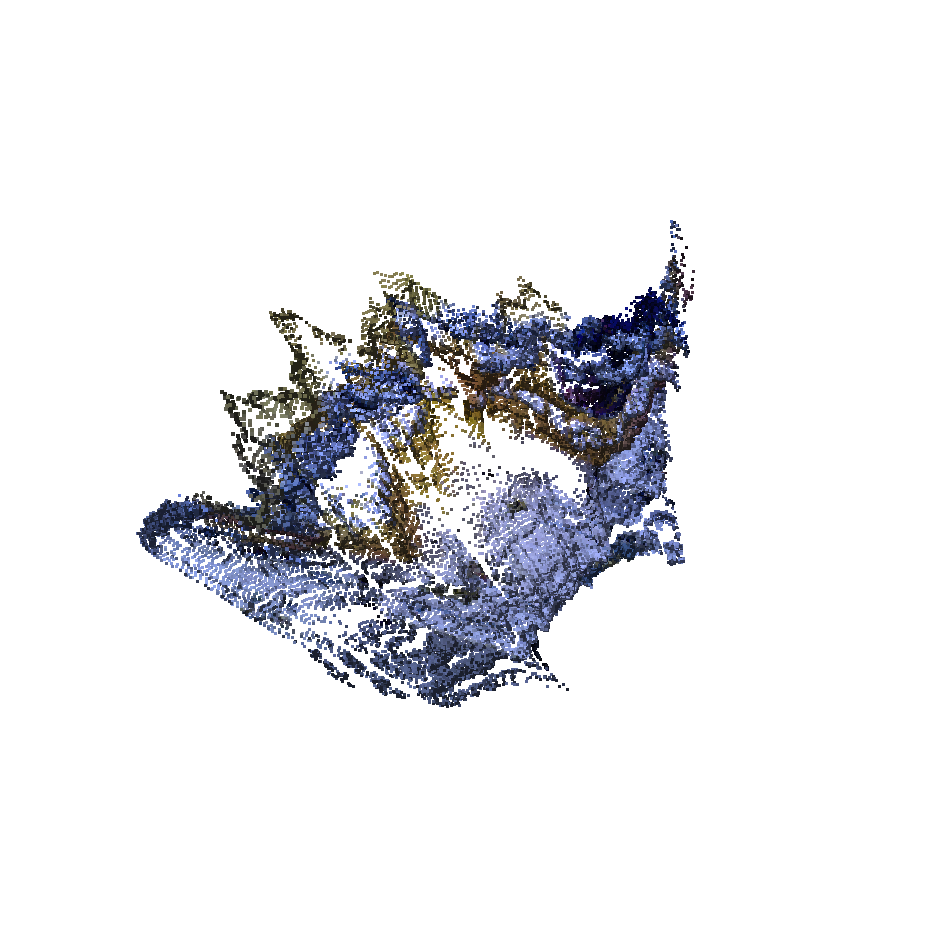}\hspace{0.5cm}
     \includegraphics[width = 1.3cm, trim = 130 130 130 130]{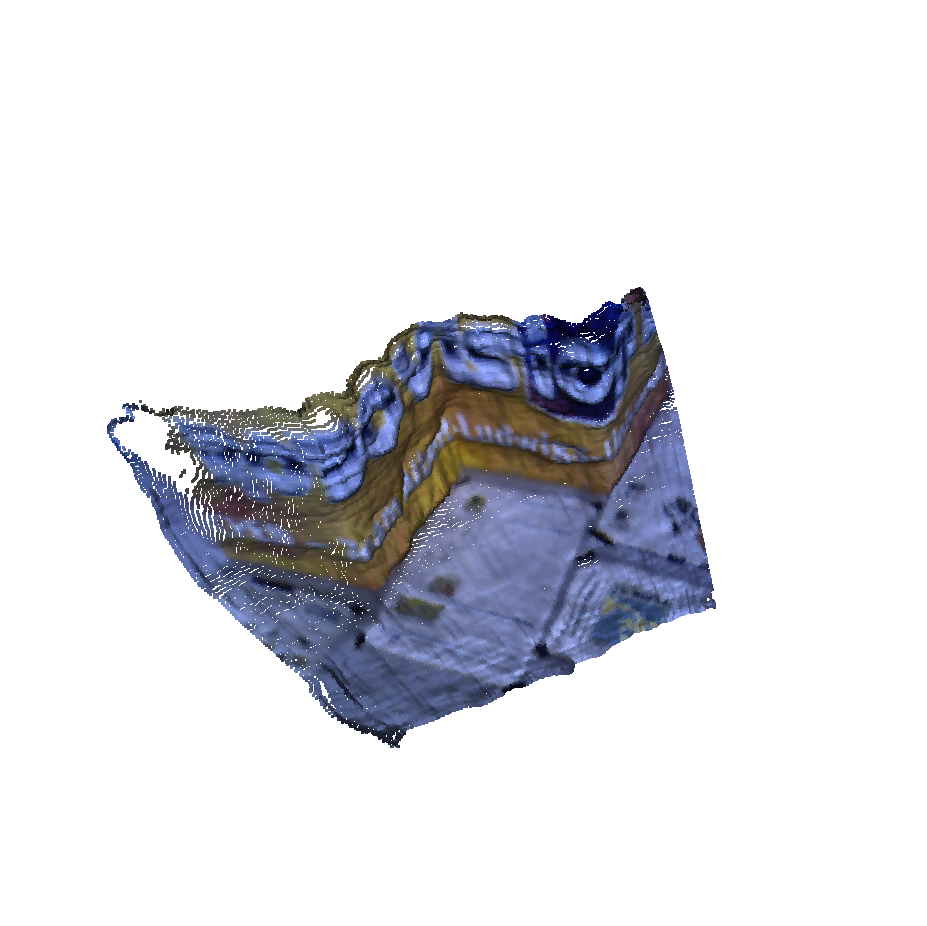}
    \\
     \includegraphics[width = 1.5cm]{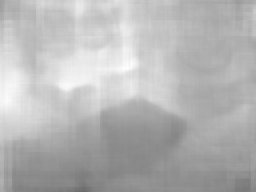}
    \includegraphics[width = 1.5cm]{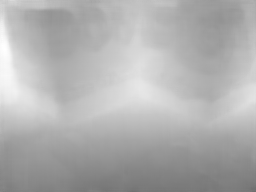}
    \includegraphics[width = 1.5cm]{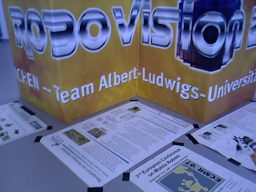}
        \\
    \hspace{0.3cm}
    \footnotesize{\textbf{Mono}} \hspace{0.4cm} \footnotesize{\textbf{Photometric}} \hspace{0.3cm} \footnotesize{\textbf{Input}}\hspace{0.4cm}
    \label{fig:my_label}
    \caption{\textbf{Dataset TUM RGB-D seq2.} }
     \end{subfigure}}\hspace{-0.3cm}
      \resizebox{0.265\textwidth}{!}{\begin{subfigure}[b]{0.3\textwidth}
    \centering
    \vspace{-0.5cm}
 \includegraphics[width = 1.9cm, trim = 150 130 150 150]{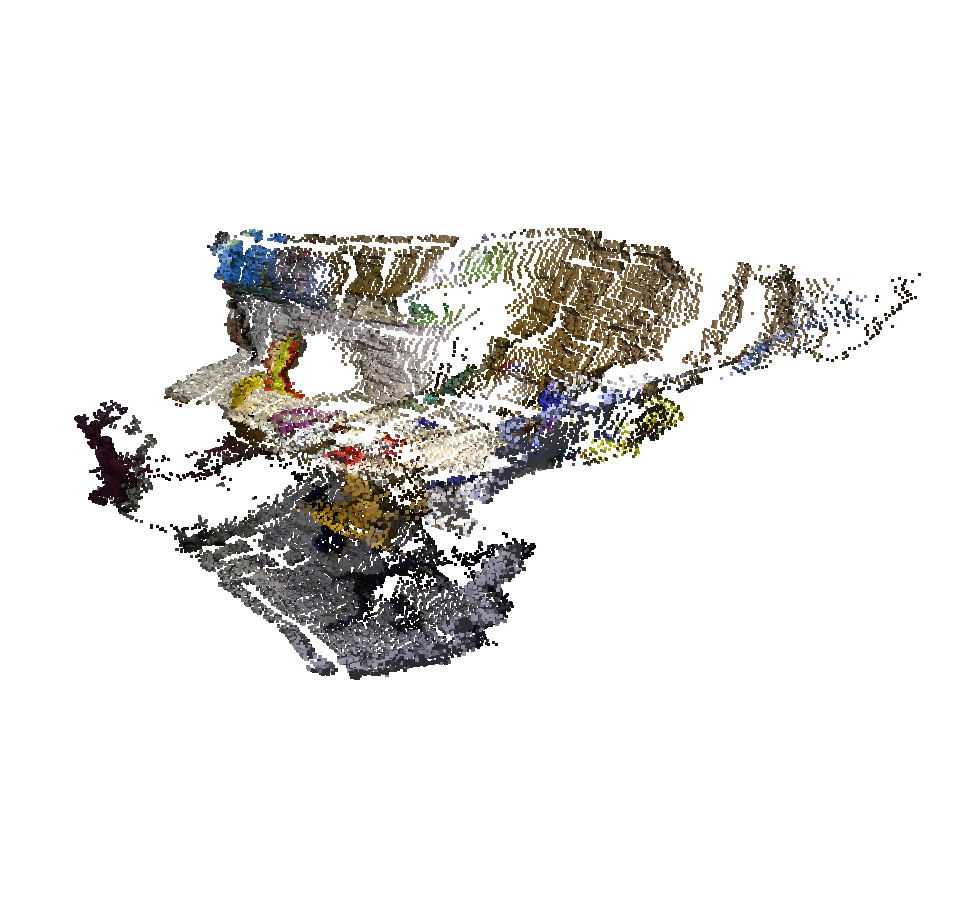}
 \hspace{0.1cm}
    \includegraphics[width = 1.9cm, trim = 80 190 80 190]{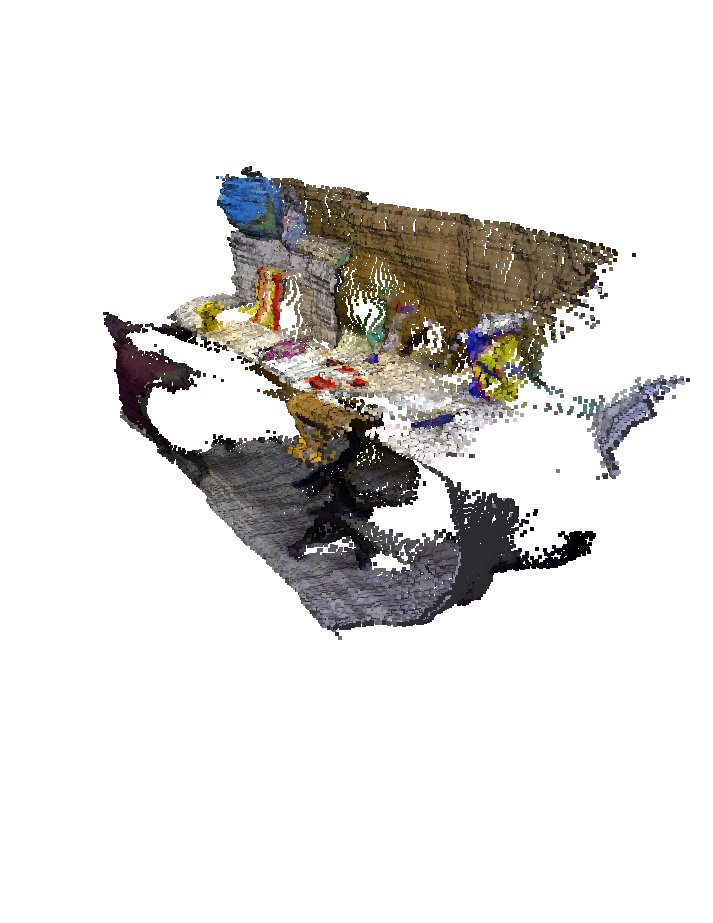}
    \\
      \includegraphics[width = 1.5cm]{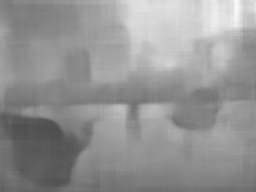}
    \includegraphics[width = 1.5cm]{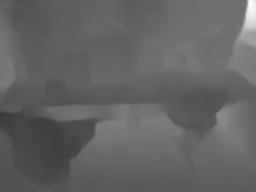}
    \includegraphics[width = 1.5cm]{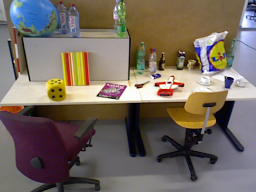}
        \\
    \hspace{0.3cm}
    \footnotesize{\textbf{Mono}} \hspace{0.4cm} \footnotesize{\textbf{Photometric}} \hspace{0.3cm} \footnotesize{\textbf{Input}}\hspace{0.4cm}
    \label{fig:my_label}
    \caption{\textbf{Dataset TUM RGB-D seq3.} }
     \end{subfigure}}\hspace{-0.4cm}
    \resizebox{0.265\textwidth}{!}{\begin{subfigure}[b]{0.3\textwidth}
    \centering
    \vspace{-0.5cm}
    \hspace{0.2cm}
 \includegraphics[width = 1.9cm, trim = 100 100 100 100]{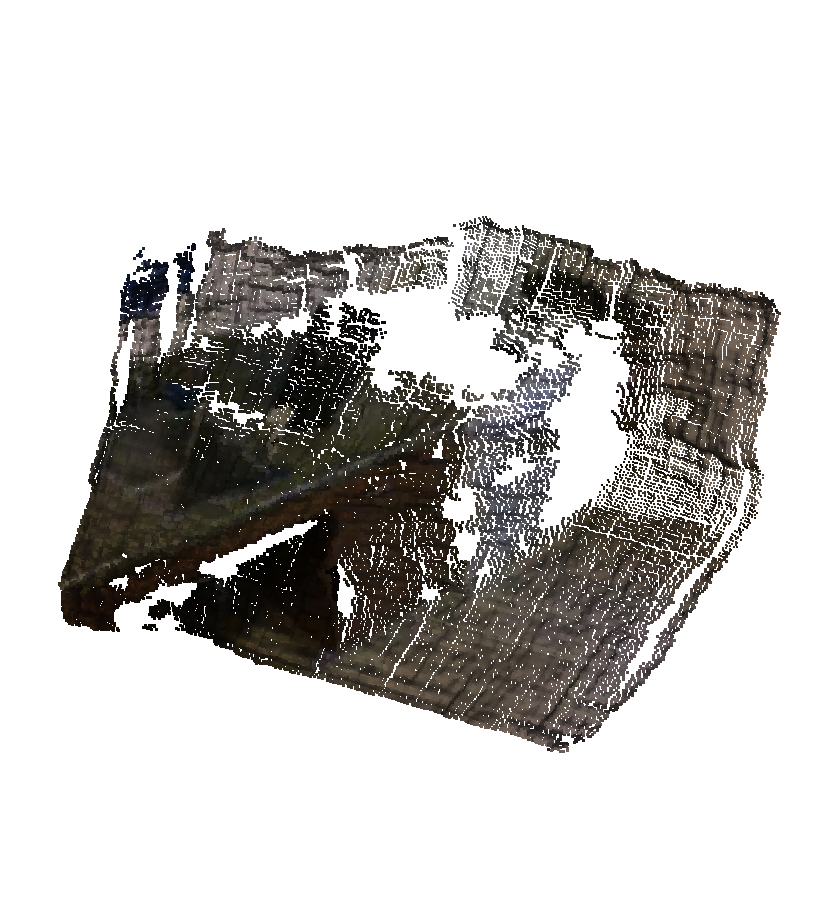}\hspace{0.4cm}
    \includegraphics[width = 1.9cm, trim = 100 100 100 100]{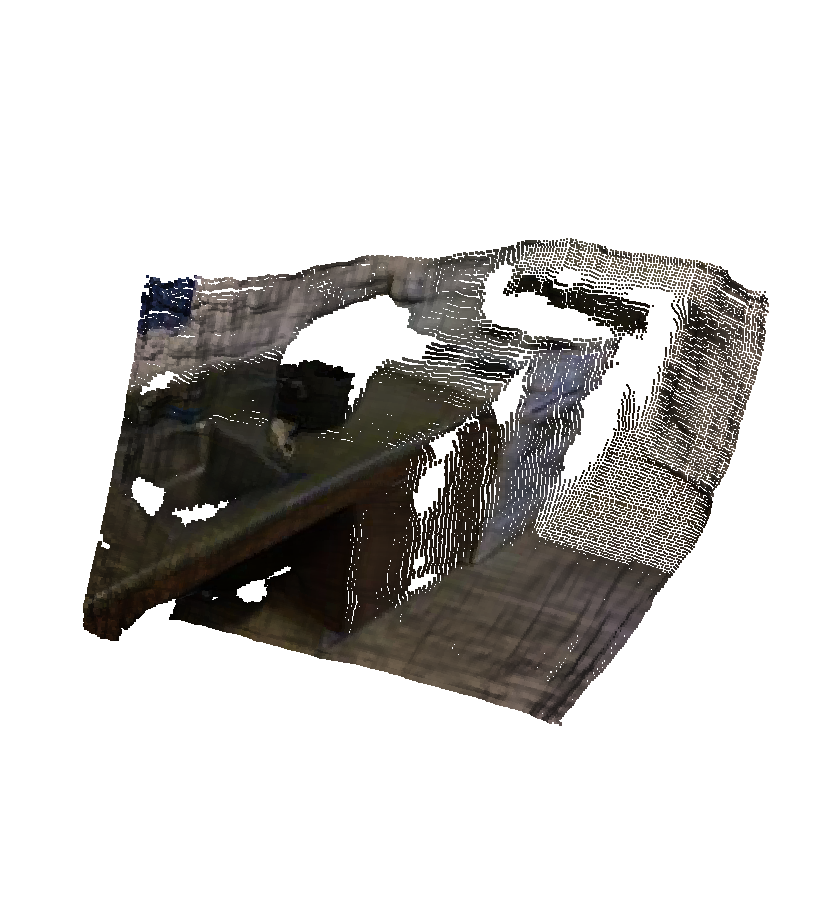}
    \\
     \includegraphics[width = 1.5cm]{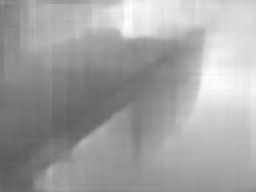}
    \includegraphics[width = 1.5cm]{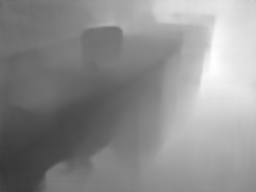}
    \includegraphics[width = 1.5cm]{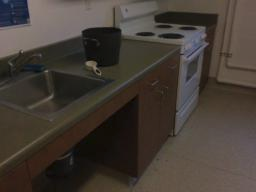}
        \\
    \hspace{0.3cm}
    \footnotesize{\textbf{Mono}} \hspace{0.4cm} \footnotesize{\textbf{Photometric}} \hspace{0.3cm} \footnotesize{\textbf{Input}}\hspace{0.4cm}
    \caption{\textbf{Dataset Scannet scene0707\_00}. }
     \end{subfigure}}\hspace{-1cm}
     \caption{\textbf{Effects of latent variables.} Examples of predicted depth maps from regular monocular depth-network (Mono) and Photometric method (with $z$ variable). Their corresponding reconstructions are shown above. }
     \label{fig:MonovsPhoto}
\end{figure*}
\begin{table}[!htbp]
\begin{center}
\resizebox{0.48\textwidth}{!}{\begin{tabular}[ht!]{|l|l|c c|c c|}
\hline
 & & \multicolumn{2}{|c|}{lower is better} & \multicolumn{2}{|c|}{higher is better}\\
Dataset & Method
&Abs Rel  & log RMSE$^{sc.}_{inv.}$ & $\delta<1.1$ & $\delta<1.25$  \\ 
\hline
\textbf{Scannet/scene0707\_00}&Monocular & 0.1590 & 0.0182& 39.79\% & 77.14\%\\ 
&Photometric & 0.0805 & 0.0071& 73.50\% & 93.44\%\\ 
\hline
\textbf{Scannet/scene0715\_00}&Monocular & 0.1097& 0.0100 & 56.27\% & 88.46\%\\ 
&Photometric & 0.0582 & 0.0045& 83.40\%&  96.67\%\\  \hline

\textbf{Scannet/scene0799\_00}&Monocular & 0.1568& 0.0197 & 46.30\% & 76.33\%  \\
&Photometric &0.0658 & 0.0056  & 79.43\% & 95.00\%  \\
\hline
\textbf{TUM/seq1}&Monocular & 0.1710& 0.2136& 35.86\% &70.61\% \\ 
&Photometric & 0.1261 & 0.0164& 72.43\% & 83.56\% \\
\hline
\textbf{TUM/seq2}&Monocular & 0.5670& 0.0256 & 32.76\% & 77.14\% \\ 
&Photometric & 0.0409& 0.0063& 90.56\% & 95.57\%\\
\hline
\textbf{TUM/seq3} &Monocular &0.2005 & 0.0491 &  38.57\% & 68.08\%\\ 
&Photometric &0.0910 &0.0187 & 74.93\% & 88.25\%\\
\hline
\end{tabular}}
\caption{Average metrics comparing ours (Photometric with $z$ variables) with regular monocular depth network.}
\vspace{0.3cm}
\label{table:monovsphotometric}
\resizebox{0.48\textwidth}{!}{\begin{tabular}[h]{|l|l|c c|c c|}
\hline
 & & \multicolumn{2}{|c|}{lower is better} & \multicolumn{2}{|c|}{higher is better}\\
Dataset & Method
&Abs Rel  &log RMSE$^{sc.}_{inv.}$ & $\delta<1.1$ & $\delta<1.25$ \\ 
\hline\hline
\textbf{Scannet/scene0707\_00}&Stereo  &  0.1330  & 0.2119 & 58.02\% & 84.79\% \\ 
&Photometric (Ours) & 0.1384 & 0.0186& 57.39\% & 82.80\%\\ 
\hline
\textbf{Scannet/scene0715\_00}&Stereo & 0.1609 & 2.2170 &  55.87\% &  78.94\% \\
&Photometric (Ours) & 0.1157 &  0.0171  & 63.75\% & 85.08\% \\
\hline
\textbf{Scannet/scene0799\_00}& Stereo& 0.1866&0.9313 &  50.48\%& 74.13\%\\ 
&Photometric (Ours)&0.1125& 0.0164& 64.48\% & 85.33\% \\
\hline
\textbf{TUM/seq1}& Stereo& 0.2832& 8.4654& 65.15\% & 76.58\% \\ 
&Photometric (Ours)& 0.1492 & 0.0212& 67.75\% & 80.20\% \\
\hline
\textbf{TUM/seq2}& Stereo & 0.1350& 7.1718& 57.82\% &  80.18\%\\ 
&Photometric (Ours)& 0.0588& 0.0095& 87.40\% & 93.43\%\\
\hline
\textbf{TUM/seq3}& Stereo & 0.1350& 7.1718& 57.82\% & 80.18\%\\ 
&Photometric (Ours) & 0.1134& 0.0282& 67.97\% &  85.25\%\\
\hline
\end{tabular}} 
\caption{Results comparing ours (Photometric with $z$ variables) using 2 co-visible cameras with stereo method in \cite{olsson-etal-cvpr-2013}.}
\label{table:stereo}
\end{center}
\end{table}

\section{Evaluation with Supervised Learning}\label{sec:supervisedtraing}
In this section we will evaluate the properties of our network architecture.
For simplicity and for the purpose of fair comparisons we will limit ourselves to a relatively standard supervised learning approach (our self supervised approach is presented in Section~\ref{sec:selfsupervisedtraining}).  
Given pairs of images $I_i$ and ground truth depths $D^{gt}_i$, we form the loss function

\begin{align}
    \mathcal{L} =  \sum_i \left( \| D_i - D^{gt}_i\|^2 + \lambda_1 \|z_i\|^2 \right) + \lambda_2 \|\textbf{W}\|^2. \label{eq:supervised}
\end{align}
Here $D_i=d( N(I_i,Q_i,z_i|\textbf{W}),\alpha_i)$ and $\alpha_i$ is calculated as the mean of the ground truth depth. The $L^2$ loss over $z$ and \textbf{W} acts as a regularizer.

The loss is minimised batch-wise  with respect to the network weights \textbf{W} and $z$ variables, using an AdaMax optimizer with a learning rate of $10^{-3}$. We train on a subset of the Scannet training datasets \cite{dai2017scannet}. The subset was drawn randomly from every second available scene/folder. In order not to obtain prohibitive training times we drew in total $150000$ samples from these scenes/folders.

\subsection{Ablation Studies}\label{sec:ablation}
Our approach for depth map estimation relies on both the network identifying a set of feasible depth maps from the image and the use of traditional stereo to pinpoint the desired solution. In the following two experiments we compare out joint model with the performance of these two components separately. The tests are run on Scannet scenes \cite{dai2017scannet} and the abbreviated TUM scenes seq1=\textit{freiburg3\_nostructure\_texture\_near\_withloop}, seq2=\textit{freiburg3\_structure\_texture\_far\_validation} and seq3= \textit{freiburg3\_long\_office\_household} \cite{sturm12iros}.\\

\noindent
\textbf{Effects of Latent Variables.} To investigate the impact of the additional latent variables we train two networks on the same Scannet dataset scenes.  The networks have the structure as described in Figure~\ref{fig:basic_network}, with the exception that in the second network, the $z$ input has been removed which turns it into a regular monocular depth predictor.
These networks were then evaluated on a subset of the test datasets of Scannet. 
For the depth parameterizing network, co-visible images were extracted (see appendix for a description), and the process described in Section \ref{sec:shape_optimization} to estimate the depth maps. 
Table \ref{table:monovsphotometric} shows metrics for the two networks. The shape parameterizing network is referred to as photometric and the monocluar depth predictor network is referred to as monodepth. Further, we show some qualitative results of the two model performances in Figure \ref{fig:MonovsPhoto}.
In all tested cases there is a substantial improvement when using photometric information.
In addition we note that the monocular network performs worse on the TUM sequences than the Scannet sequences. This is an effect of training on Scannet (albeit on different scenes) and poor generalization properties.
It is clear from the results that the ability to use photometric information makes our approach generalize much better to the TUM sequences.  

\begin{figure}
\vspace{0.2cm}
\resizebox{0.475\textwidth}{!}{
    \includegraphics[width = 18mm]{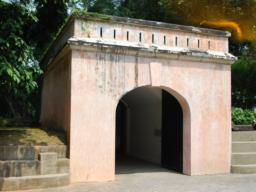}
       \includegraphics[width = 18mm]{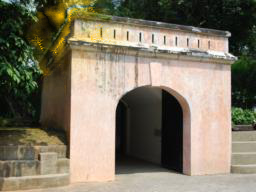}
       \includegraphics[width = 18mm]{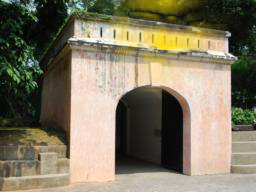}
       \includegraphics[width = 18mm]{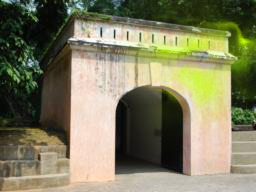}
       \includegraphics[width = 18mm]{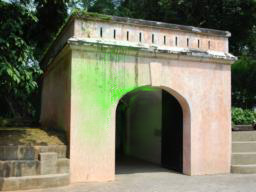}
          \includegraphics[width = 18mm]{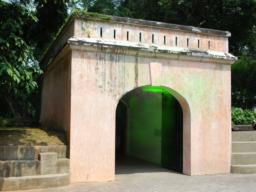}
       }
       \\
       \resizebox{0.47\textwidth}{!}{ \includegraphics[width = 18mm]{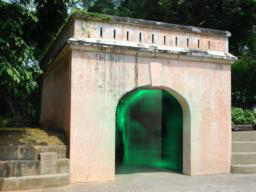}
       \includegraphics[width = 18mm]{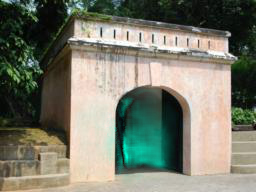}
   \includegraphics[width = 18mm]{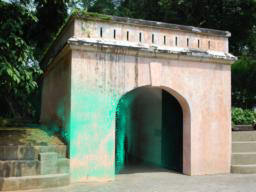}
           \includegraphics[width = 18mm]{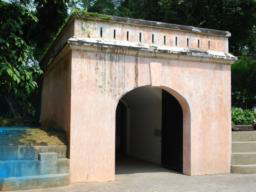}
       \includegraphics[width = 18mm]{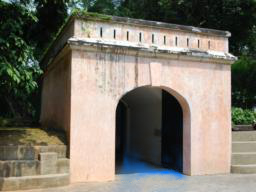}
       \includegraphics[width = 18mm]{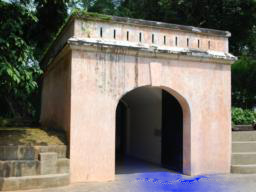}}
       \caption{\textbf{Local influence of $z$-variable}. Each figure shows the local effect of a small shift in one dimension of the $z$-variable. The change is coded in color.  }
    \label{fig:z_heatmap}
\end{figure}

Figure~\ref{fig:z_heatmap} displays the impact of the components in $z$ on the estimated depth. From the figure we can see that the components have a local impact, with a structure and size that depends on the image. See appendix for more examples.
\noindent
\textbf{Stereo Comparison.} 
To evaluate the contribution of the learning component we next compare our approach with a publicly available stereo method \cite{olsson-etal-cvpr-2013} relying on photometric information. This method uses rectified cameras and therefore we limit our method to use only pairs of images for a fair comparison. Due to the arbitrary camera motion of the dataset, only a subset of the the pairs are suitable for rectification. We therefore filter out unsuitable pairs that have few remaining valid pixel after rectification. The results are shown in Table \ref{table:stereo}. Note that our formulation generally achieves better results than the stereo method despite only optimizing over 192 variables. This suggests that the shape parameterization that is provided by our network is able to reduce the dimensionallity of the search-space without any significant loss of accuracy. Further, the stereo method tend to produce unrealistic representations of shape due to the fitted tangent planes, see appendix for more details.

\section{Multiple View Stereo for Self-Supervision}\label{sec:selfsupervisedtraining}

In this section we present the self-supervised training. Using this approach we can train the network end-to-end without ground truth depth maps, in exactly the way it is expected to be used later. 
Our approach trains the network to learn the mapping $ \rho = N(I,Q,z)$ that maximizes
\begin{align}
\prod_i p(\mathcal{I}_{-i}, \mathcal{D}_{-i}|D_i,I_{i}).
\end{align}
Here $D_i = d(N(I_i,Q_i,z_i | \textbf{W}), \alpha_i)$, $\mathcal{I}_{-i}$ is the set of all images co-visible with $I_i$ except $I_i$ itself and $\mathcal{D}_{-i}$ defined similarly. Notice here that we do not assume that we know the ground truth value of $\mathcal{D}_{-i}$. Rather these are current estimates of the depth in the co-visible views. When we update the weights \textbf{W} and the latent variables $z_i$, we always update co-visible images jointly. We assume that the photometric and depth variables are conditionally independent and can be factorized as
\begin{align}
   p(\mathcal{I}_{-i}, \mathcal{D}_{-i}|D_i,I_{i}) =  p(\mathcal{I}_{-i}|D_i,I_{i})p(\mathcal{D}_{-i}|D_i,I_{i})
\end{align}
Under the assumption that the distributions belong to the exponential family, we can define the negative log-likelihood of the photometric and depth distribution as in Equations \eqref{eq:lphoto} and \eqref{eq:ldepth}. The normalizing constants do not depend on the network weights \textbf{W} and $z_i$ and are therefor neglected. To this we add the additional loss 
\begin{align}
\mathcal{L}_{\alpha_i} = \left( 1- \frac{1}{N} \sum_p \frac{1-\rho_{i}(p)}{\rho_{i}(p)} \right)^2
\end{align}
motivated by
$
\alpha_i = \frac{1}{N} \sum_p D_{i}(p) = \frac{\alpha_i}{N} \sum_p \frac{1-\rho_{i}(p)}{\rho_{i}(p)}
$
which only holds if $\mathcal{L}_{\alpha_i} = 0$. The loss will penalize solutions where $\alpha_i$ is not the mean of the current depth map. The total loss becomes
\begin{align}
\mathcal{L} = \sum_i (  \mathcal{L}_{photo_i} + \mathcal{L}_{depth_i} + \mathcal{L}_{\alpha_i} + \lambda_1 \| z_i\|^2) + \lambda_2 \|\textbf{W}\|^2   
\label{eq:trainingloss}
\end{align}
We minimize this loss batch-wise w.r.t. \textbf{W}, $\alpha_{i}$ and $z_i$, for ${i=1,...,M}$ using an AdaMax optimizer with learning rate of $10^{-3}$. 

When experimentally comparing training with ground truth depth maps from Section~\ref{sec:supervisedtraing} with the self supervised approach (using the same number of images from Scannet training dataset) we observe that both methods yield similar results (see appendix for an evaluation). This shows that the latter option is viable and furthermore we could expect to see an improvement if more training data was used as indicated in \cite{garg2016unsupervised,MegaDepthLi18,monodepth17, monodepth2}. 

\begin{table*}[ht!]
\begin{center}
\resizebox{\textwidth}{0.45\textwidth}{\begin{tabular}{|l|l|c c c c|c c c|}
\hline
 & & \multicolumn{4}{|c|}{lower is better} & \multicolumn{3}{|c|}{higher is better}\\
Dataset/scene & Method
&Abs Rel & Sq Rel  & RMSE & log RMSE$^{sc.}_{inv.}$ & $\delta < 1.1$ & $\delta < 1.25$& $\delta < 1.25^2$\\ 
\hline\hline
\textbf{Scannet/scene0565\_00}&DeepFactors & 0.1517 & 0.0693& 0.3638& 0.0202& 46.08\%& 76.68\%& 95.67\%\\ 
&MegaDepth & 0.2749 & 0.2879 & 0.6672&0.0577 &36.46\% & 62.34\%& 83.20\% \\ 
&PSMNet & 0.3272 & 0.5718& 0.6209& 0.0803 & 48.28\% & 73.03\%& 85.30\%\\ 
&Ours  & \textbf{0.0980}&\textbf{0.0492}& \textbf{0.3036}& \textbf{0.0105} & \textbf{67.31\%} & \textbf{88.70\%}&\textbf{97.83\%} \\ 
\hline
\textbf{Scannet/scene0606\_02}&DeepFactors & 0.1736 & 0.1546& 0.5799&0.5173 &44.96\%&73.61\% & 91.01\% \\ 
&MegaDepth & 0.2312 & 0.1998 &0.5959 &0.0401 &35.70\% &63.30\% & 87.39\%\\ 
&PSMNet & 0.1523 & 0.1599 & 0.4648 & 0.0393 & 59.52\% &82.71\% & 91.92\% \\ 
&Ours  & \textbf{0.1232}& \textbf{0.0804}& \textbf{0.3989} & \textbf{0.0193} &  \textbf{61.62\%}& \textbf{83.73\%}& \textbf{95.32\%}\\  \hline
\textbf{Scannet/scene0707\_00}&DeepFactors & 0.1669& 0.0913& 0.3771&0.0234 &43.02\%&73.62\% & 94.70\%\\ 
&MegaDepth  & 0.2452 & 0.2226& 0.5443& 0.0463& 33.36\%& 61.90\%& 86.73\% \\ 
&PSMNet & 0.2384 & 0.3275& 0.5929&0.0691 &45.65\% & 70.39\% &  83.68\%\\ 
&Ours  & \textbf{0.0674}& \textbf{0.0369}& \textbf{0.2530}&\textbf{0.0065}&  \textbf{79.42\%}&\textbf{94.69\% }& \textbf{98.68\%}\\ \hline
\textbf{Scannet/scene0715\_00}&DeepFactors & 0.0959& 0.0653 &0.4599&0.0199 &64.16\%&90.03\% & 97.74\%\\ 
&MegaDepth  & 0.2291 & 0.4771& 0.9298& 0.0413 & 45.54\% &73.12\% & 89.03\% \\ 
&PSMNet &  0.1799 & 0.2068 & 0.6944 & 0.0540 & 49.28\% &75.20\% & 87.28\% \\ 
&Ours  & \textbf{0.0674}& \textbf{0.0408}& \textbf{0.3355}&\textbf{0.0065}& \textbf{79.42\%}&\textbf{94.69\% }& \textbf{98.68\%}\\ 
\hline
\textbf{Scannet/scene0743\_00}&DeepFactors & 0.1537& 0.0607& 0.751& 0.2340& 45.61\%&78.26\% & 95.70\%\\  
&MegaDepth  & 0.2111& 0.1236& 0.3570  & 0.0380 & 39.86\%& 70.00\% &  89.74\%\\  
&PSMNet & 0.1852 & 0.1595 & 0.4783 & 0.0602 & 47.46\% & 76.04\% & 89.01\% \\ 
&Ours  & \textbf{0.0823} & \textbf{0.0020}&\textbf{0.1640}& \textbf{0.0073} & \textbf{72.01\%}& \textbf{92.60\%}& \textbf{98.68\%}\\  \hline
\textbf{Scannet/scene0799\_00}& DeepFactors & 0.1537& 0.1999& 0.6070&0.02751& 45.61\%&78.26\% & 95.70\%\\ 
&MegaDepth & 0.1929& 0.2096& 0.6937& 0.0300 & 44.91\%& 73.72\% & 91.05\%\\  
&PSMNet & 0.1559 & 0.1877& 0.7622& 0.0452 & 56.06\% & 78.70\% & 90.12\% \\ 
&Ours  & \textbf{0.0808} & \textbf{0.0548}& \textbf{0.4230}& \textbf{0.0088 }& \textbf{73.52\%}&\textbf{91.35\%} & \textbf{98.02\%}\\ \hline
\textbf{TUM/fr3\_structure\_texture\_far\_validation}&DeepFactors & 0.1166&0.0744&0.4922& 0.0152&56.35\%& 85.18\% &96.37\% \\
&MegaDepth  & 0.2186& 0.3806&0.8846&0.0405 &42.28\%& 70.55\%& 90.44\% \\ 
&PSMNet & 0.1096 & 0.4055& 0.7234& 11.72 & 81.90\% & 92.29\% & 95.27\% \\ 
&Ours & \textbf{0.0497} & \textbf{0.0440}&\textbf{0.3955}&\textbf{0.0084} & \textbf{88.49\%}& \textbf{92.84\%}& \textbf{97.51\%} \\ \hline
\textbf{TUM/fr3\_structure\_texture\_far}&DeepFactors & 0.1054 &0.0647&0.4690&0.0127 & 60.74\%&87.46\% &96.71\% \\ 
&MegaDepth  & 0.2169& 0.3076& 0.7830 & 0.0411 & 41.13\%& 69.73\% & 90.32\%  \\ 
&PSMNet &0.2027  & 0.5990& 0.8483& 14.13& 39.53\% & 73.31\% & 93.60\% \\ 
&Ours & \textbf{0.0318} & \textbf{0.0256}& \textbf{0.3416}&  \textbf{0.0043} &  \textbf{93.84\%}& \textbf{96.77\%}& \textbf{98.52\%}\\  \hline
\textbf{TUM/fr3\_nostructure\_texture\_near\_withloop}&DeepFactors & 0.1464 & 0.0839&\textbf{0.2307}& 0.9937 & 43.94\%&79.12\% & \textbf{97.72\%}\\  
&MegaDepth  & \textbf{0.1370}&\textbf{0.0468}&0.2344 &\textbf{0.0162}&47.58\% &81.29\% & 96.94\%\\ 
&PSMNet & 0.2189 & 1.1133& 0.6378& 10.4871 & \textbf{78.25\%} & \textbf{87.00\%} & 91.07\% \\ 
& Ours & 0.1869 &0.1334& 0.3364 & 0.0285 &\ 65.38\%&74.57\% &88.06\% \\  \hline
\textbf{TUM/fr3\_long\_office\_household} & DeepFactors & 0.1656 &0.1830&0.8120& 0.0424& 48.01\%& 76.27\%& 89.62\%\\ 
&MegaDepth & 0.2433 & 0.3099& 0.9397& 0.0601 & 34.20\% & 62.97\% & 83.97\% \\ 
&PSMNet & 0.1528 & 0.2665& 0.8550& 22.33 & 66.61\% &  83.80\%& 90.16\%\\ 
&Ours & \textbf{0.1030} &  \textbf{0.0970}& \textbf{0.5755}&\textbf{0.0248 }& \textbf{71.89\%}& \textbf{86.06\%}& \textbf{93.32\%}\\  \hline
\textbf{TUM/fr1\_xyz} & DeepFactors &0.1155 &0.0669&0.2789&0.2908 &  57.13\%& 86.61\%& 97.47\%\\ 
&MegaDepth  & 0.2922& 0.3399& 0.5321&0.0622 &36.79\%& 67.71\%& 87.38\%\\ 
&PSMNet & 01722 & 0.1313 & 0.3974 & 29.38 & 66.84\% & 81.48\%  & 89.17\% \\ 
&Ours &\textbf{ 0.0881}&\textbf{ 0.0264}&  \textbf{0.2060}& \textbf{0.0097}& \textbf{72.95\%}& \textbf{88.26\%}&\textbf{98.29\%} \\  \hline
\textbf{TUM/fr2\_rpy}& DeepFactors & 0.2396 & 0.5624 & 1.0491&5.0664& 42.88\%& 67.91\%& 81.99\%\\ 
&MegaDepth  & 0.2783& 0.3347&1.0024&0.0831&32.25\% &57.43\% & 75.25\% \\  
&PSMNet & 0.2895 & 0.4457& 1.0305& 24.38& 41.19\% & 63.90\% & 76.08\% \\ 
&Ours & \textbf{0.1766 }&\textbf{0.1924}& \textbf{0.7896}& \textbf{0.0415}& \textbf{54.89\%}&\textbf{73.21\% }&\textbf{84.65\%} \\  \hline
\end{tabular}}
\caption{\textbf{Quantitative results.} Evaluation of our system on validation scenes from Scannet \cite{dai2017scannet} and TUM RGB-D\cite{sturm12iros}. We compare our system against DeepFactors \cite{Czarnowski:2020:10.1109/lra.2020.2965415}, MegaDepth \cite{li2018megadepth} and PSMNet \cite{chang2018pyramid}. All metrics are presented as averages of all samples. }
\label{table:othermethods}
\end{center}
\end{table*}
\begin{figure*}[ht!]
    \centering
    \resizebox{\textwidth}{!}{
    \includegraphics[width = 18mm]{Images/Comparing_all/images/1341839268304930_rgb.png}
 \includegraphics[width = 18mm]{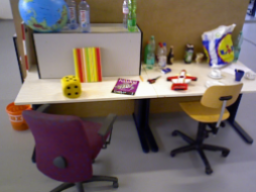}
 \includegraphics[width = 18mm]{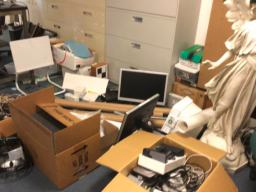}
 \includegraphics[width = 18mm]{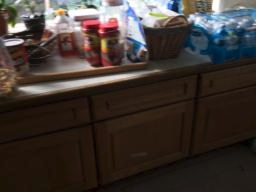}
 \includegraphics[width = 18mm]{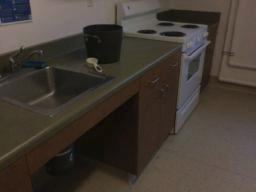}
 \includegraphics[width = 18mm]{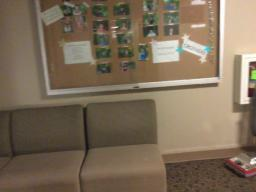}
    \includegraphics[width = 18mm]{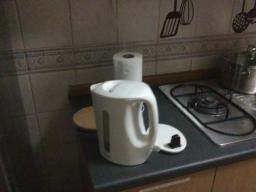}
  \includegraphics[width = 18mm]{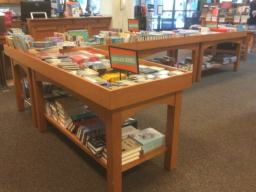}
   }\\
    \vspace{-0.5cm}
     \resizebox{\textwidth}{!}{\hspace{0.3cm}\alert{\textbf{TUM/seq2 }}\hspace{2.6cm} \alert{\textbf{TUM/seq3 } }\hspace{2.6cm}  \alert{\textbf{scannet/scene0565\_00 }} \hspace{1cm} \alert{scannet/scene0606\_02 } \hspace{1cm} \alert{scannet/scene0707\_00 } \hspace{1cm}\alert{scannet/scene0715\_00 }\hspace{1cm} \alert{scannet/scene0743\_00 } 
     \hspace{1cm} \alert{scannet/scene0799\_00 }\hspace{1cm}}
   \\
   \vspace{0.5mm}
   \resizebox{\textwidth}{!}{
    \includegraphics[width = 18mm]{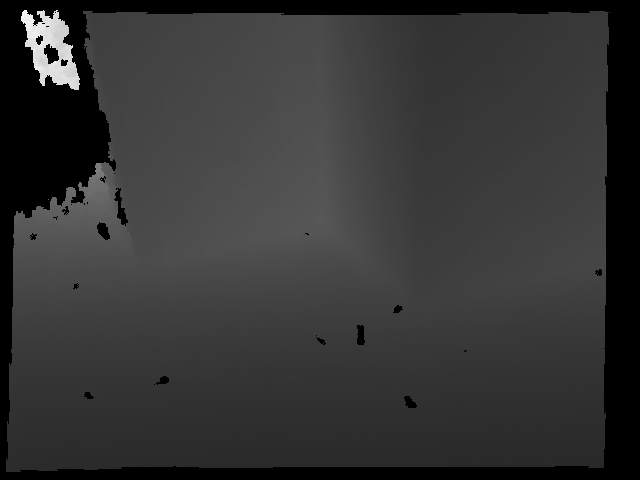}
  \includegraphics[width = 18mm]{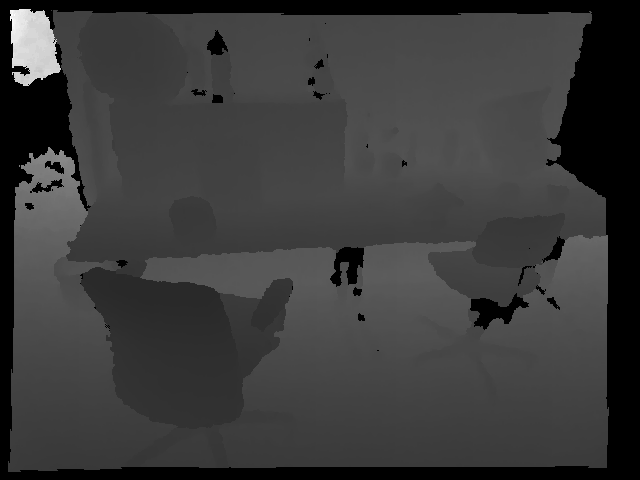}
   \includegraphics[width = 18mm]{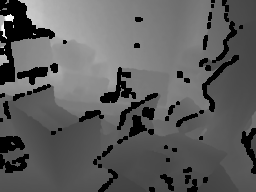}
     \includegraphics[width = 18mm]{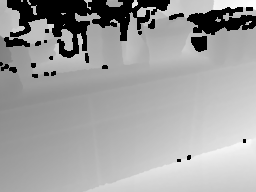}
       \includegraphics[width = 18mm]{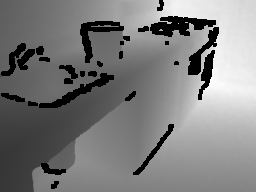}
         \includegraphics[width = 18mm]{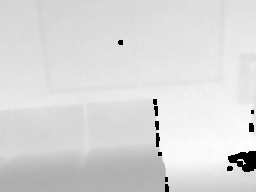}
           \includegraphics[width = 18mm]{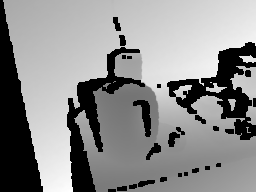}
     \includegraphics[width = 18mm]{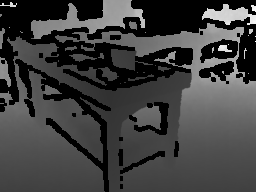}
   }\\ \resizebox{\textwidth}{!}{\includegraphics[width = 18mm]{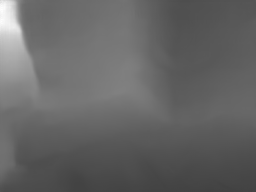}
  \includegraphics[width = 18mm]{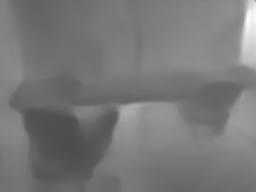}
   \includegraphics[width = 18mm]{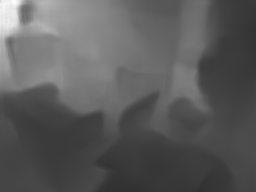}
  \includegraphics[width = 18mm]{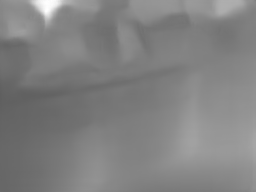}
  \includegraphics[width = 18mm]{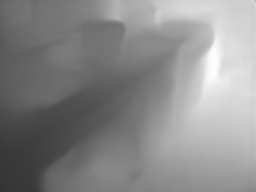}
     \includegraphics[width = 18mm]{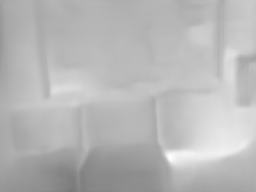}
       \includegraphics[width = 18mm]{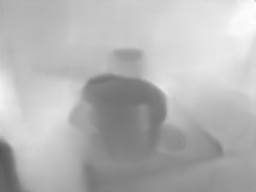}
  \includegraphics[width = 18mm]{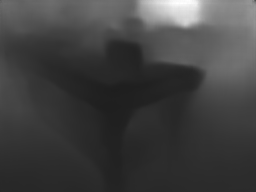}
  }
   \\ \resizebox{\textwidth}{!}{\includegraphics[width = 18mm]{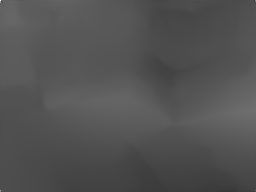}
  \includegraphics[width = 18mm]{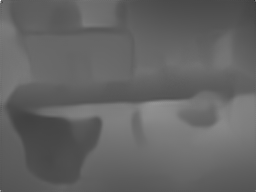}
  \includegraphics[width = 18mm]{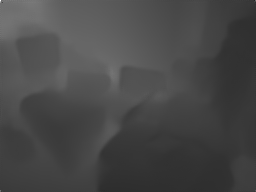}
  \includegraphics[width = 18mm]{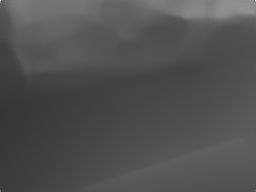}
  \includegraphics[width = 18mm]{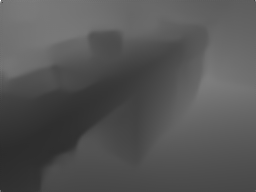}
     \includegraphics[width = 18mm]{Images/Comparing_all/Deepfactors/frame-000115.png}
       \includegraphics[width = 18mm]{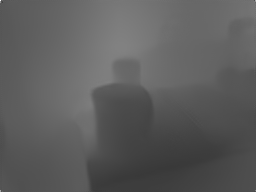}
           \includegraphics[width = 18mm]{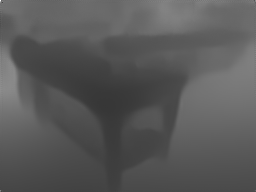}
  }
   \\ \resizebox{\textwidth}{!}{\includegraphics[width = 18mm]{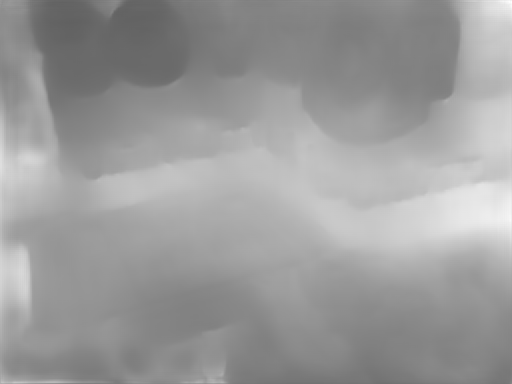}
   \includegraphics[width = 18mm]{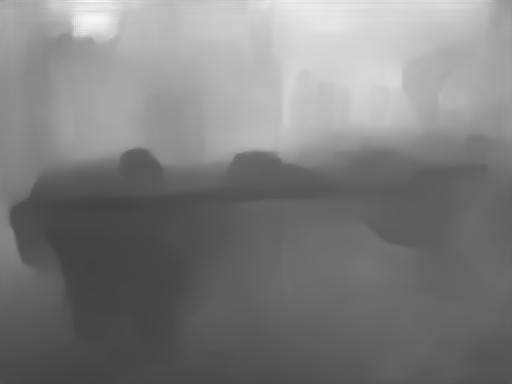}
   \includegraphics[width = 18mm]{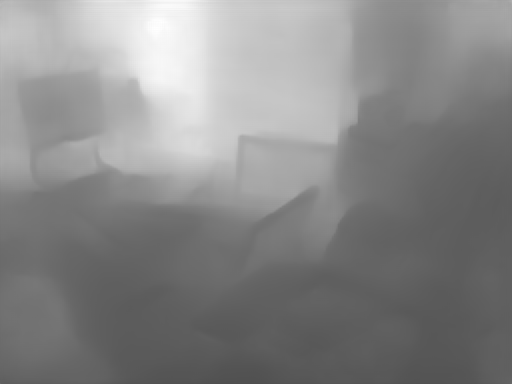}
\includegraphics[width = 18mm]{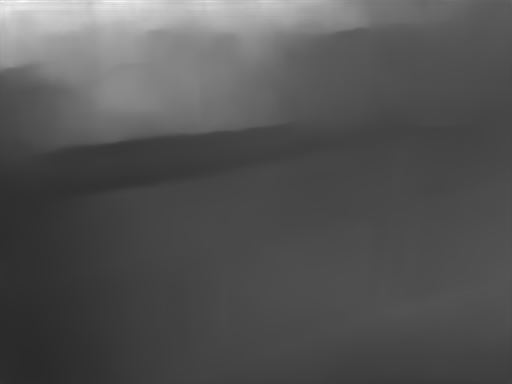}
 \includegraphics[width = 18mm]{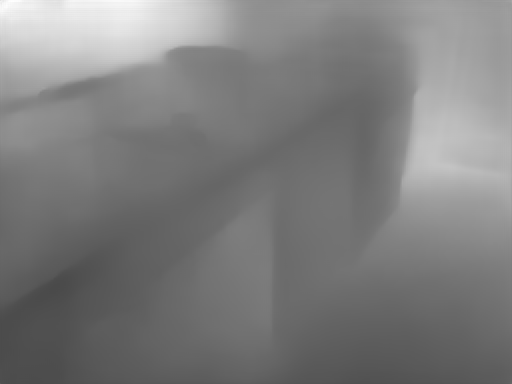}
  \includegraphics[width = 18mm]{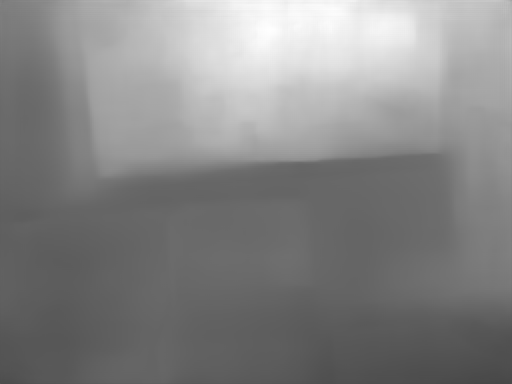}
 \includegraphics[width = 18mm]{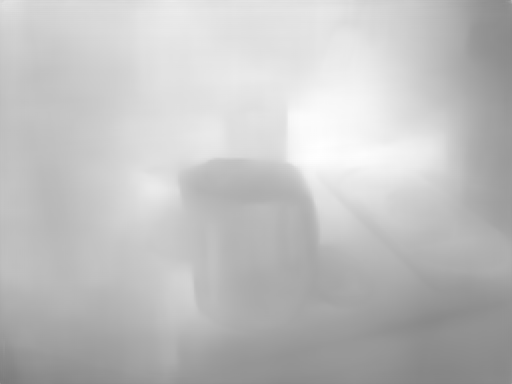}
 \includegraphics[width = 18mm]{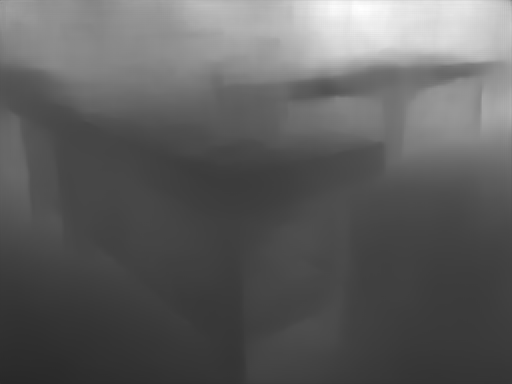}      
 }
   \\ \resizebox{\textwidth}{!}{\includegraphics[width = 18mm]{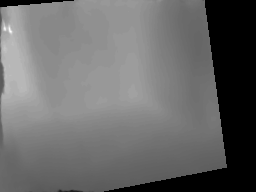}
   \includegraphics[width = 18mm]{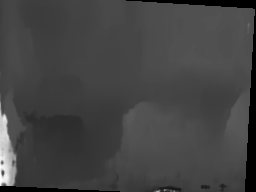}
   \includegraphics[width = 18mm]{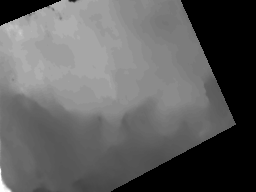}
   \includegraphics[width = 18mm]{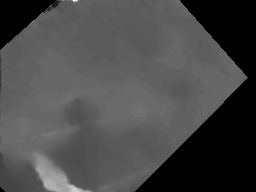}
   \includegraphics[width = 18mm]{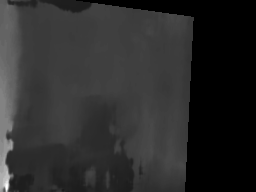}
   \includegraphics[width = 18mm]{Images/Comparing_all/psnet/stereo_frame-000760.png}
   \includegraphics[width = 18mm]{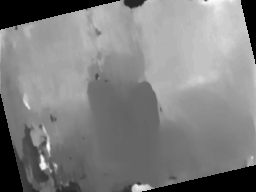}
   \includegraphics[width = 18mm]{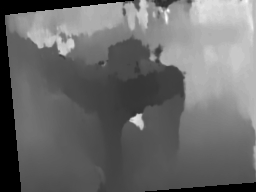}
 }\\ 
    \caption{\textbf{Qualitative results. Depth map comparison.} From the top; input image, ground truth, Ours, DeepFactors, MegaDepth and PSMNet predictions.}
    \label{fig:quantitative_othermethods}
    \resizebox{\textwidth}{!}{
     \includegraphics[width = 18mm, trim = 100 150 100 100]{Images/3D_comp_all/Ours/134183926984482800.png}
 \includegraphics[width = 18mm, trim = 100 150 100 100]{Images/3D_comp_all/DF/134183926984482800.png}
 \includegraphics[width = 18mm, trim = 100 150 100 100]{Images/3D_comp_all/MD/134183926984482800.png}
 \includegraphics[width = 18mm, trim = 80 150 80 80]{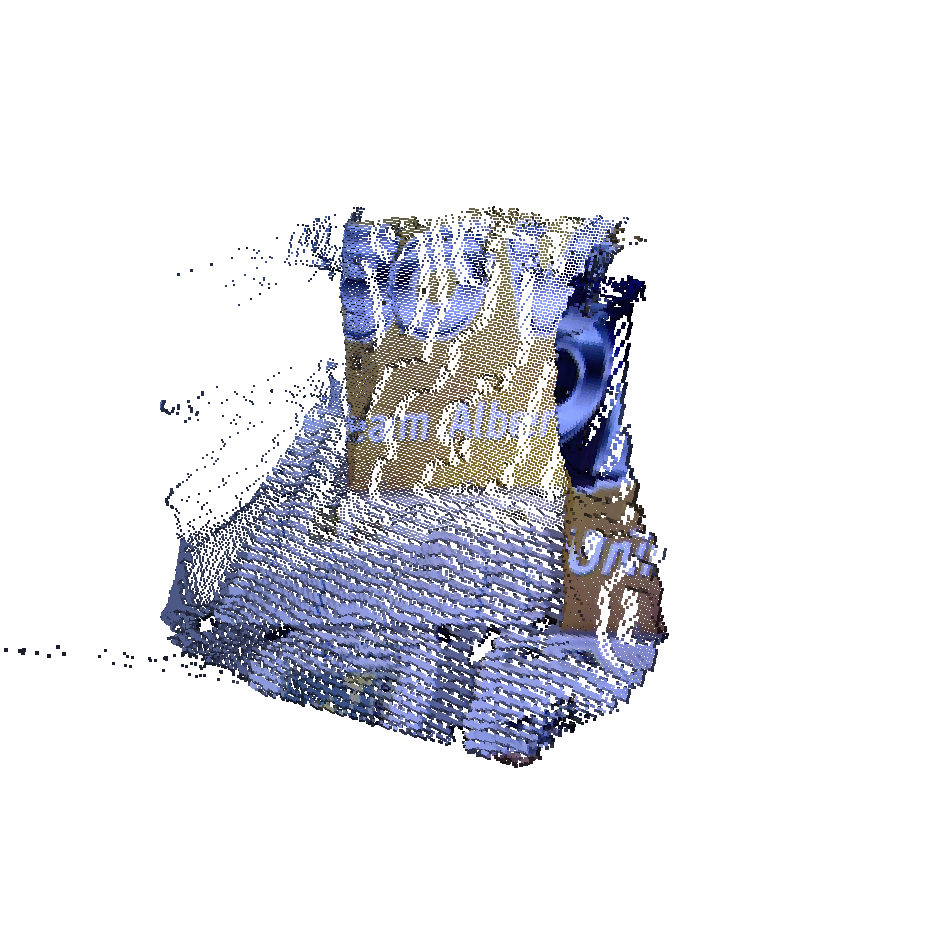}
\includegraphics[width = 16mm]{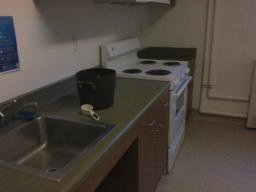}
 \includegraphics[width = 20mm, trim = 200 250 200 200]{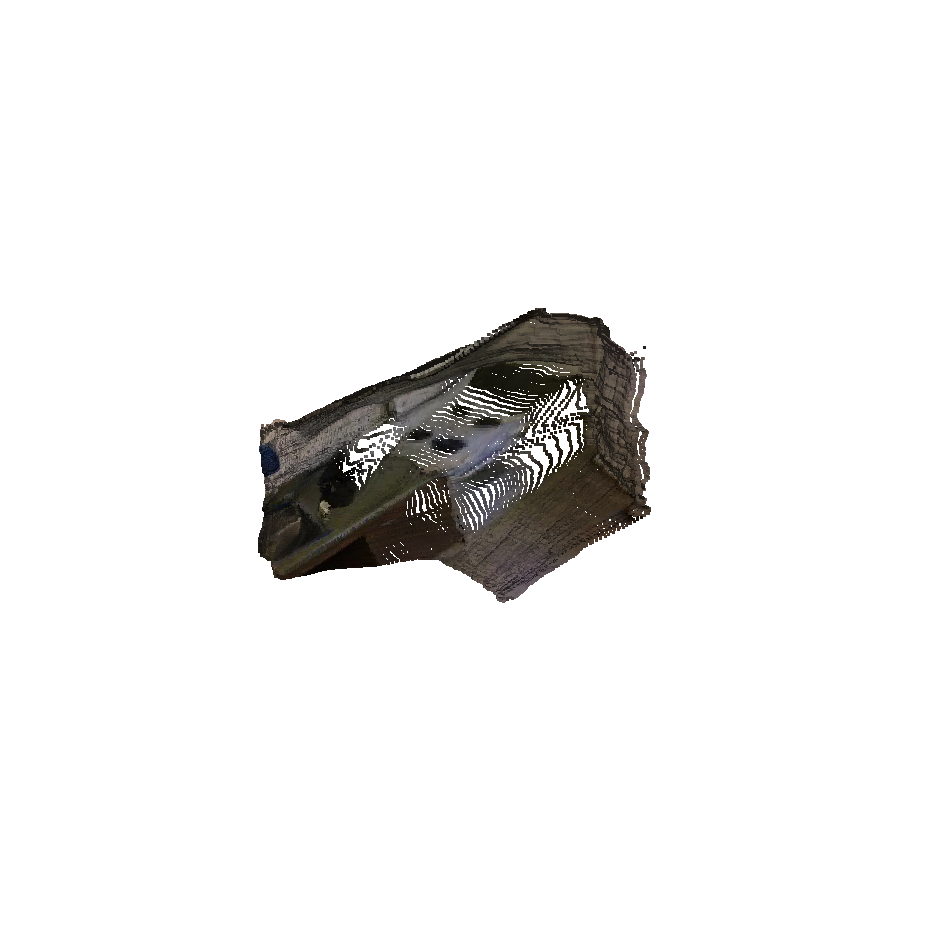}
 \includegraphics[width = 18mm, trim = 200 200 200 200]{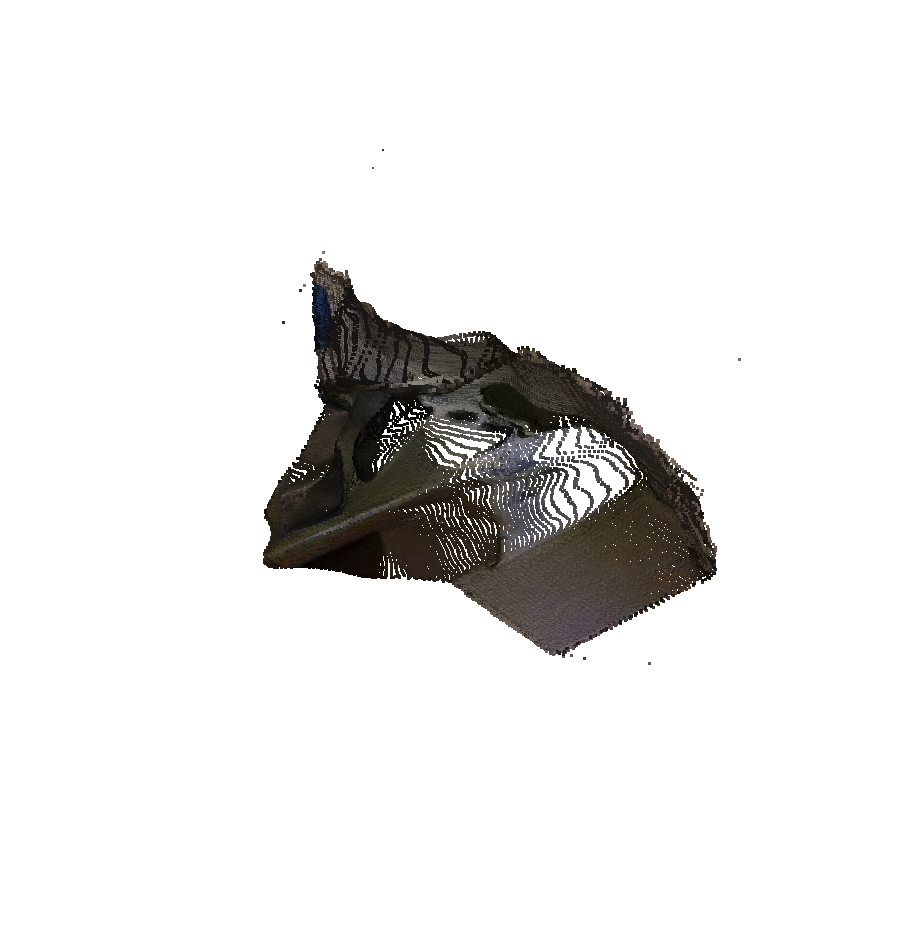}
 \includegraphics[width = 18mm, trim = 100 200 100 100]{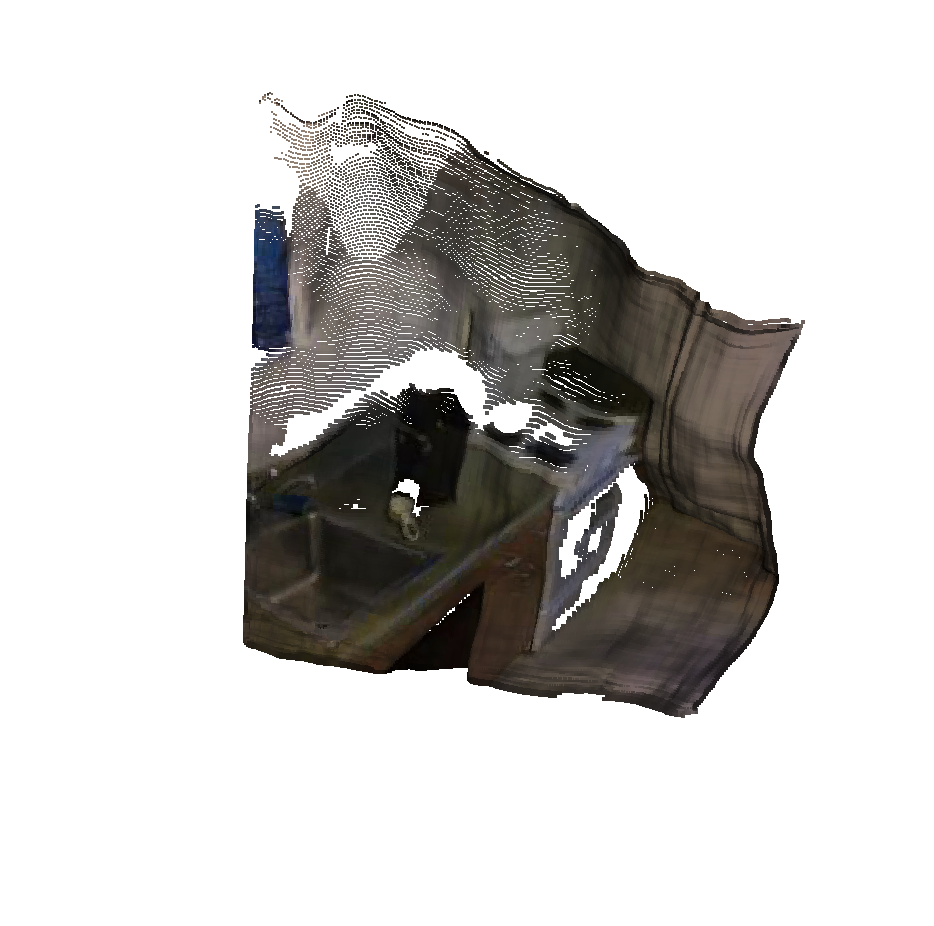}}\\
   \vspace{-0.1cm}
   \resizebox{\textwidth}{!}{\hspace{2cm}\footnotesize{\textbf{Ours}}\hspace{2cm}\footnotesize{\textbf{DeepFactors}}
   \hspace{2cm}\footnotesize{\textbf{MegaDepth}} \hspace{2cm}\footnotesize{\textbf{PSMNet}}\hspace{2cm}\footnotesize{\textbf{Image}}\hspace{2cm}
   \footnotesize{\textbf{Ours}}\hspace{2cm}\footnotesize{\textbf{DeepFactors}}
   \hspace{2cm}\footnotesize{\textbf{MegaDepth}} }
   \resizebox{\textwidth}{!}{
    \includegraphics[width = 18mm, trim = 0 200 0 300]{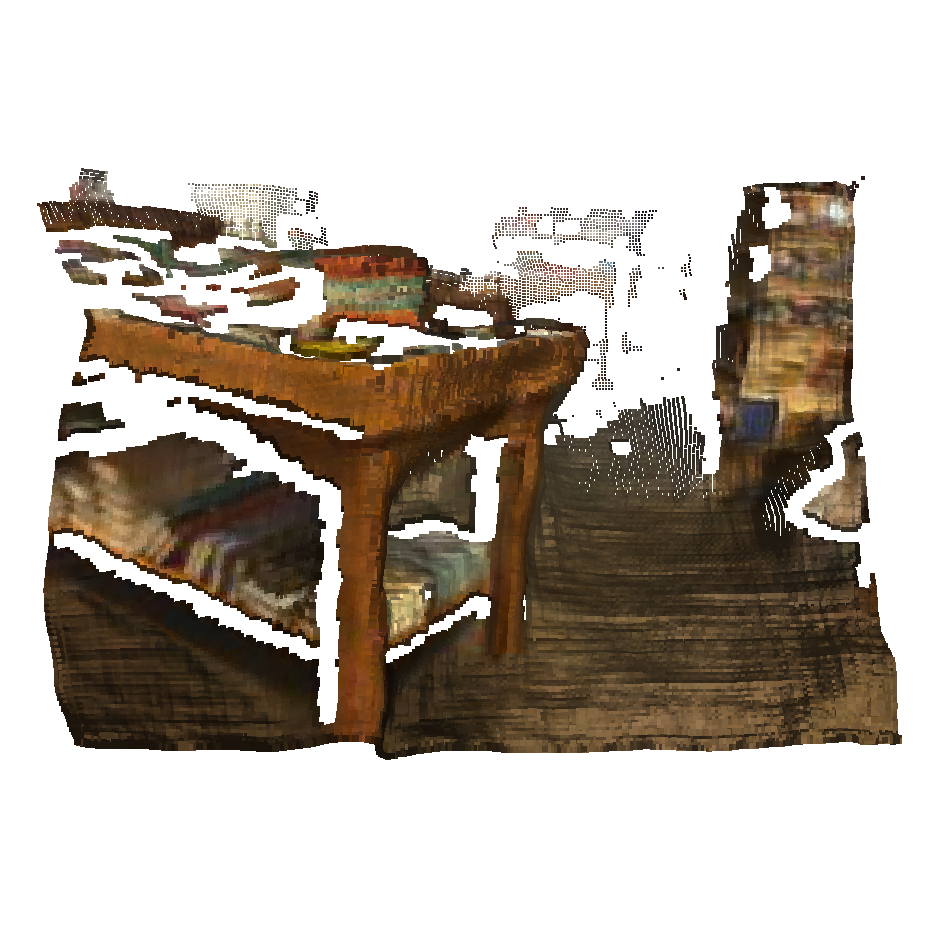}
 \includegraphics[width = 19mm, trim = 0 200 0 300]{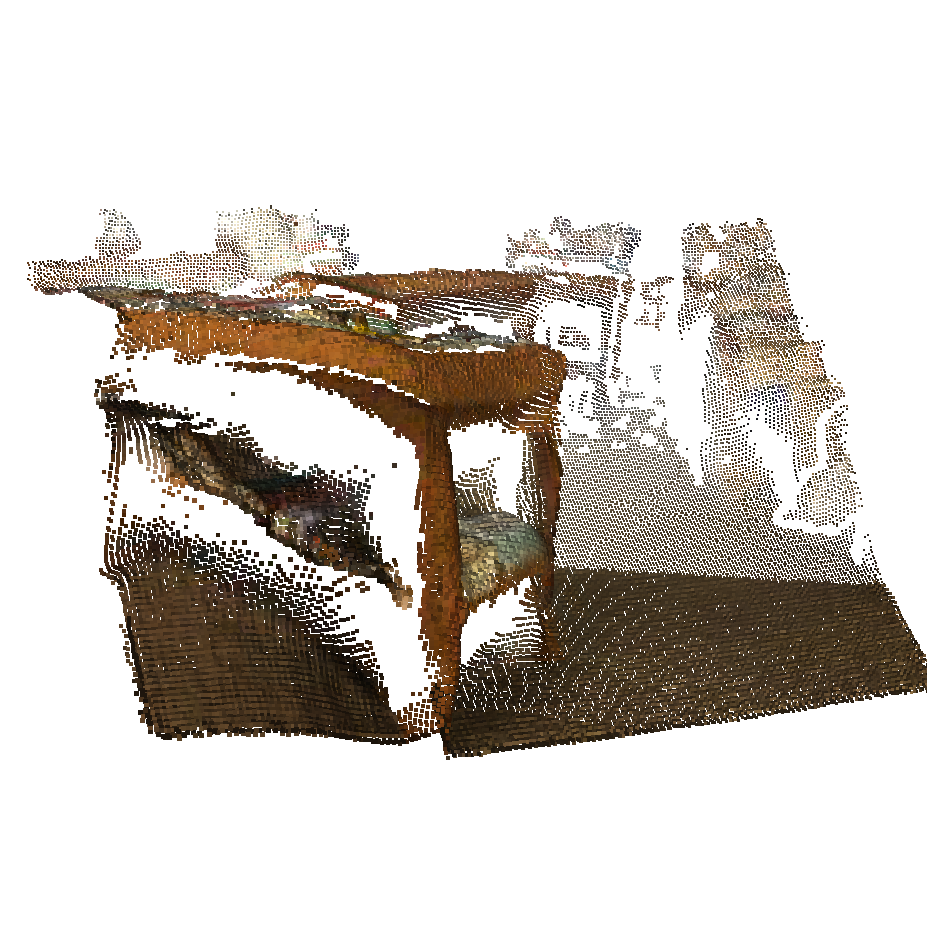}
  \includegraphics[width = 19mm, trim = 0 200 0 300]{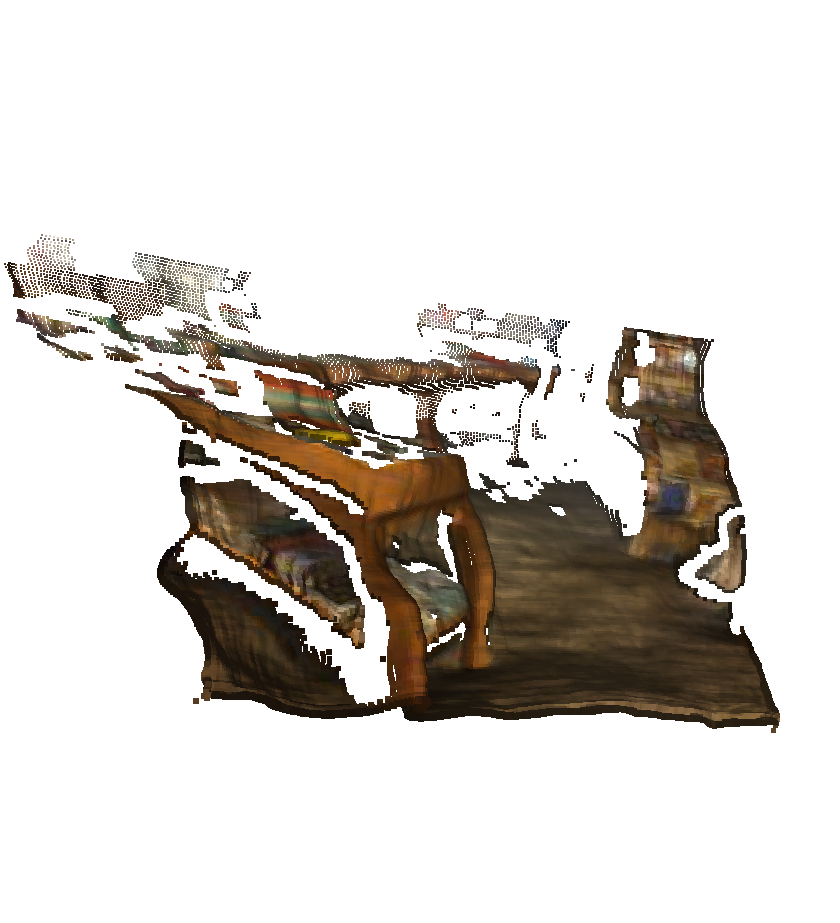}
   \includegraphics[width = 18mm, trim = 0 200 0 300]{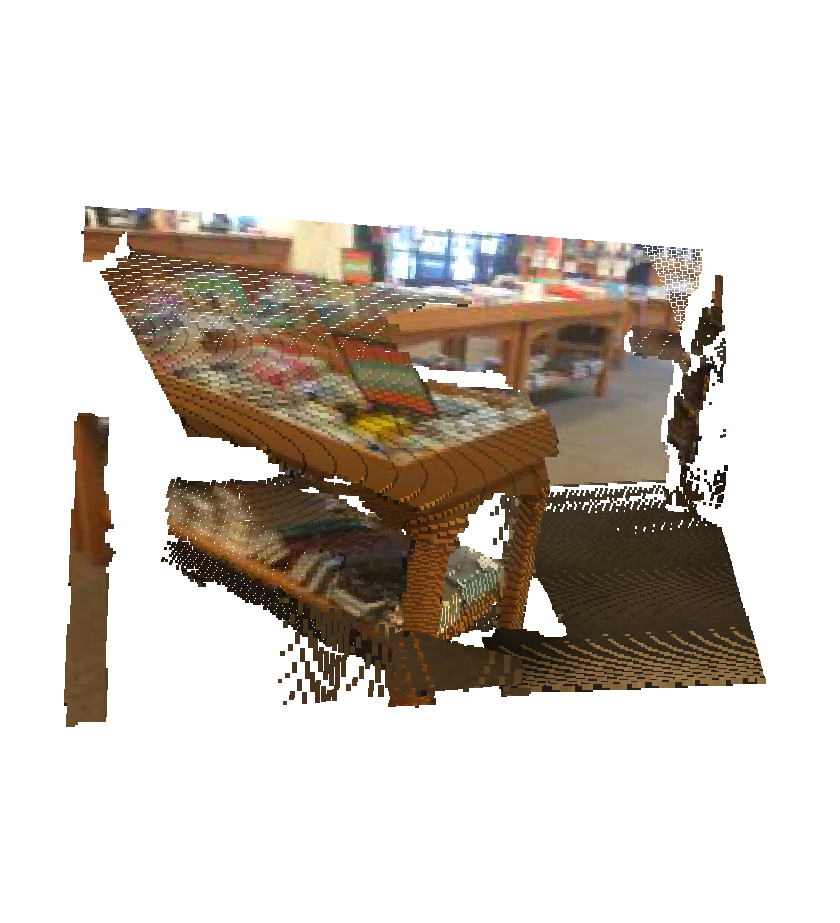}
\includegraphics[width = 16mm]{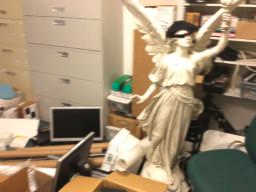}
 \includegraphics[width = 20mm, trim = 100 200 100 120]{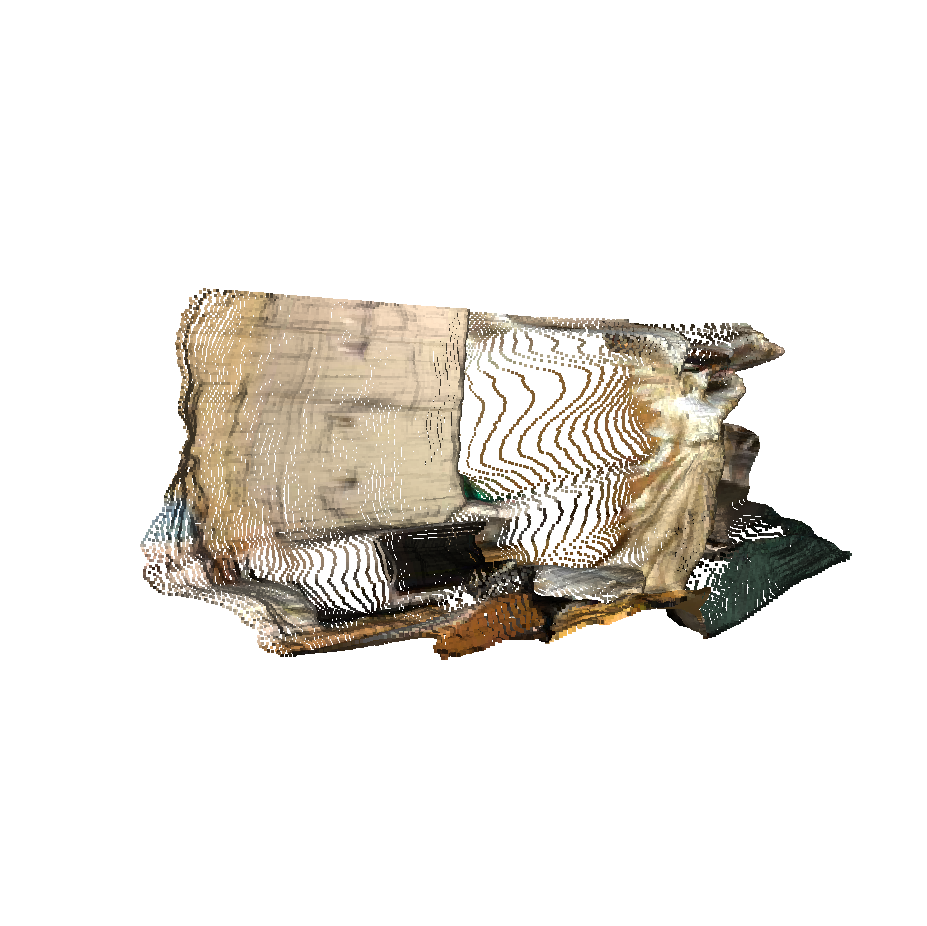}
 \includegraphics[width = 18mm, trim = 70 200 100 100]{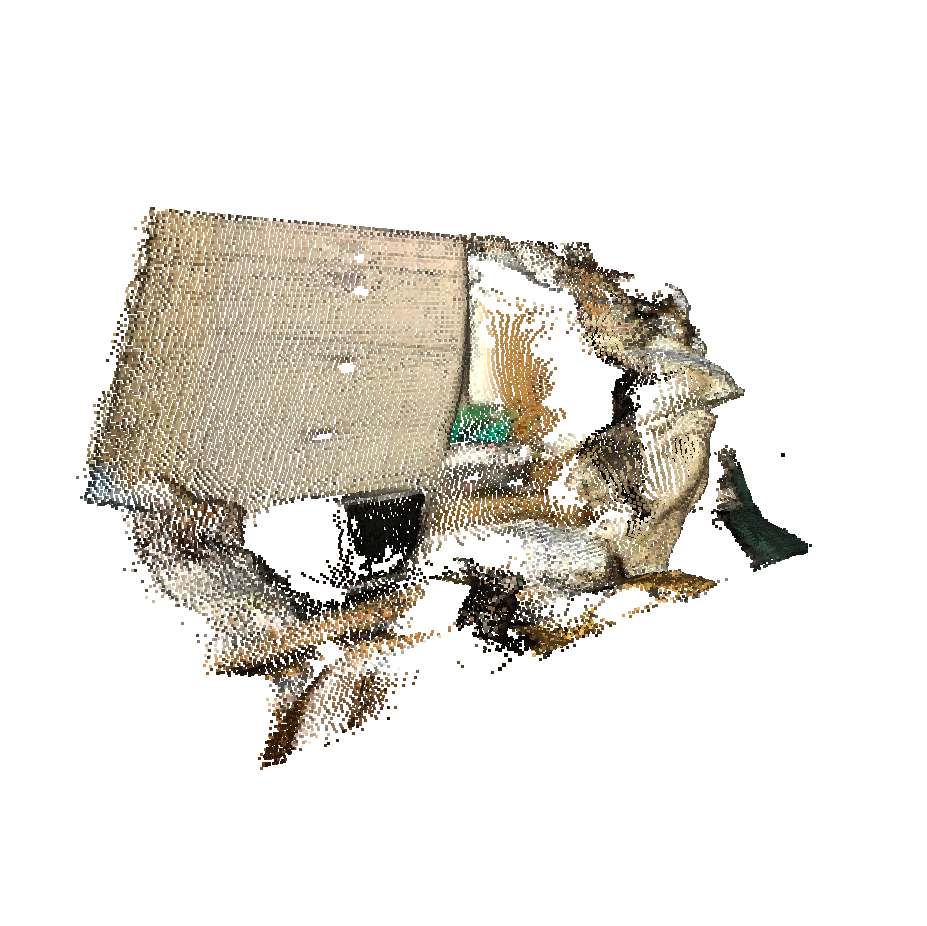}
 \includegraphics[width = 18mm, trim = 100 300 100 100]{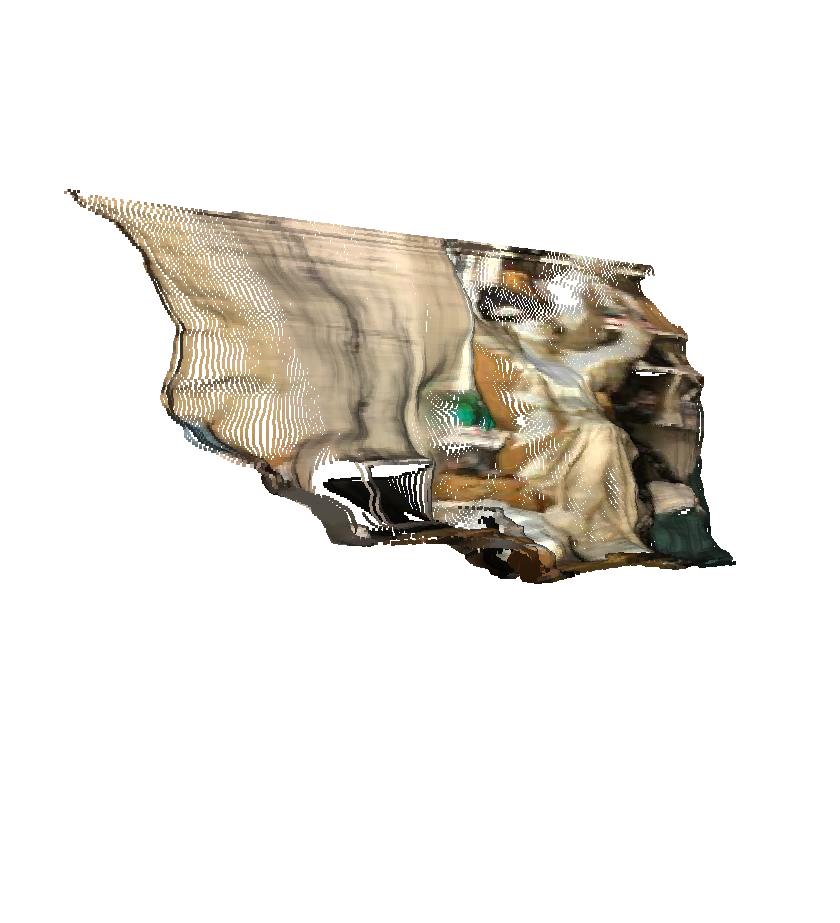}}\\
  \vspace{0.2cm}
   \resizebox{\textwidth}{!}{\hspace{2cm}\footnotesize{\textbf{Ours}}\hspace{2cm}\footnotesize{\textbf{DeepFactors}}
   \hspace{2cm}\footnotesize{\textbf{MegaDepth}} \hspace{2cm}\footnotesize{\textbf{PSMNet}}\hspace{2.5cm}\footnotesize{\textbf{Image}}\hspace{2cm}
  \footnotesize{\textbf{Ours}}\hspace{2cm}\footnotesize{\textbf{DeepFactors}} \hspace{2cm}\footnotesize{\textbf{MegaDepth} }}
    \caption{\textbf{Qualitative results.} Examples of reconstructed scenes from datasets TUM/seq2, Scannet scene0707\_00, Scannet scene0799\_00 and Scannet scene0565\_00. }
\label{fig:quantitative_othermethods3dd}
\end{figure*}

\section{Experimental Comparisons} \label{sec:comparison_to_others}
We compare our method with depth predictions from DeepFactors\cite{Czarnowski:2020:10.1109/lra.2020.2965415}, MegaDepth\cite{MegaDepthLi18} and PSMNet\cite{chang2018pyramid}. 
 For each method the predictions are scaled with the median of the fractions between predicted and ground truth depth, more specifically  $s=\textbf{Median}(D_{gt}/D_{pred})$, for a fair comparison. A similar scale is used in e.g. \cite{Zhou_2017} and \cite{Czarnowski:2020:10.1109/lra.2020.2965415}.

 The presented metrics are computed on different subsets of the dataset for the different methods. Scannet scenes are evaluated as follows; we run DeepFactors on the full dataset and calculate the metrics on the keyframes it is preconfigured to generate (to not affect the performance). MegaDepth and our method consider a larger set of the dataset. How this is selected from the testset is described in the appendix. PSMNet requires rectified images, and thus it is evaluated on the set of meaningful rectifications of the dataset.
 
 On the TUM scenes our, MegaDepth and DeepFactors are all evaluated on the keyframes selected by DeepFactors for a fair comparison. For PSMNet we evaluate on the subset of images with meaningful rectifications. 
 We trained our network on a subset of Scannet training data (v2) along with some scenes from \cite{enqvist-etal-omnivs-2011}
, using the loss function given in \eqref{eq:trainingloss}. Full description of training data is given in appendix.
The methods are subsequently evaluated on Scannet testset (v1 and v2) and TUM RGB-D datasets. 
The TUM RGB-D depths and images are associated as in \cite{sturm12iros}. We evaluate the predictions on several metrics from previous work, as RMSE, Absolute Relative difference, Squared relative difference, scale-invariant RMSE and Accuracy under a threshold \cite{10.1109/CVPR.2014.19} as in \cite{10555529690332969091, NIPS2014_5539}. The results shown in Table~\ref{table:othermethods} are means over the number of samples.
Figure~\ref{fig:quantitative_othermethods} shows a subset of the estimated depth maps. In Figure~\ref{fig:quantitative_othermethods3dd} we also plot these in 3D.

\section{Conclusions}
In this work we presented a learning approach for monocular depth estimation that takes ambiguities into account by providing a low dimensional parameterization of a family of feasible depth maps. We have shown that optimizing over this representation using photo-consistency losses yields accurate and realistic geometries. Our experimental results indicate, both qualitative and quantitative, that our approach generalizes better than competing state-of-the-art methods.

{\small
\bibliographystyle{ieee_fullname}
\bibliography{egbib}
}
	
\newpage

\twocolumn[
\begin{center}
{\Large \bf Appendix:
Monocular Depth Parameterizing Networks\\}
\end{center}
\vspace{0.8cm}
]
\appendix
\section{Matrix Model Continued}\label{app:matrix_probl}
\begin{figure*}
	\begin{center}
		\def\w{30mm}
		\includegraphics[width=\w]{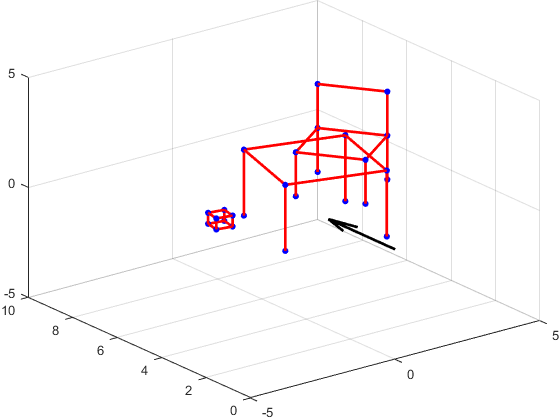}
		\includegraphics[width=\w]{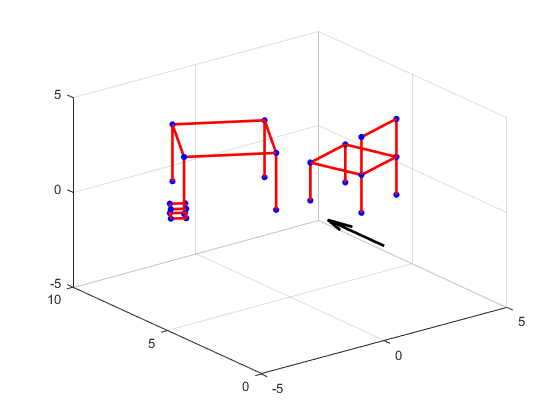}
		\includegraphics[width=\w]{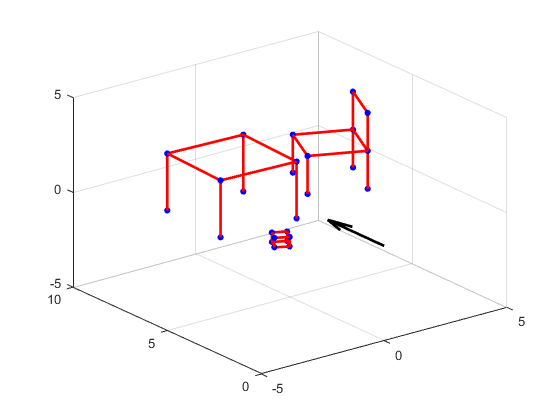}
		\includegraphics[width=\w]{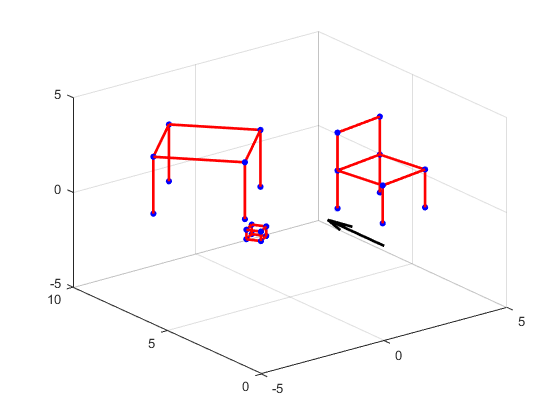}\\
		
		\includegraphics[width=\w]{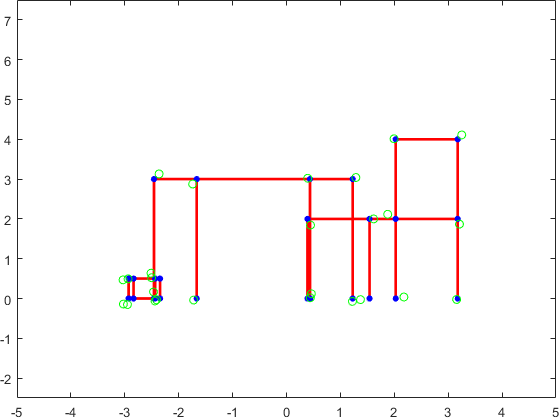}
		\includegraphics[width=\w]{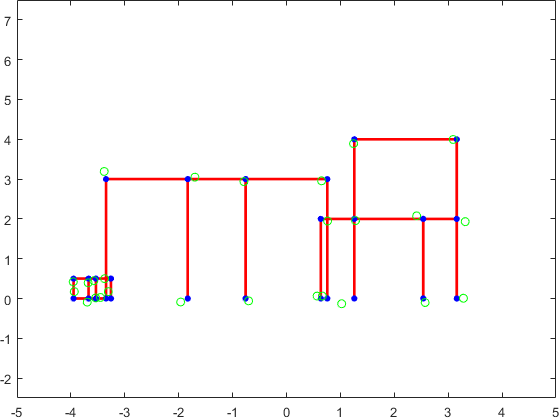}
		\includegraphics[width=\w]{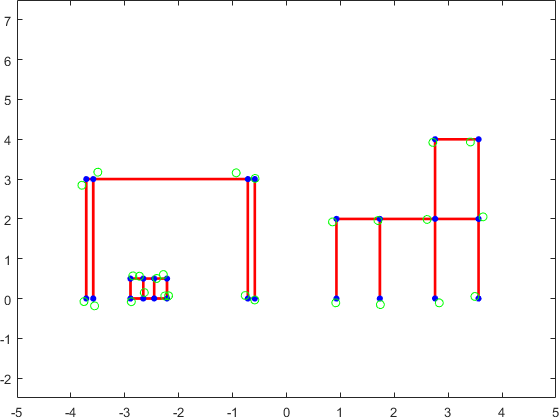}
		\includegraphics[width=\w]{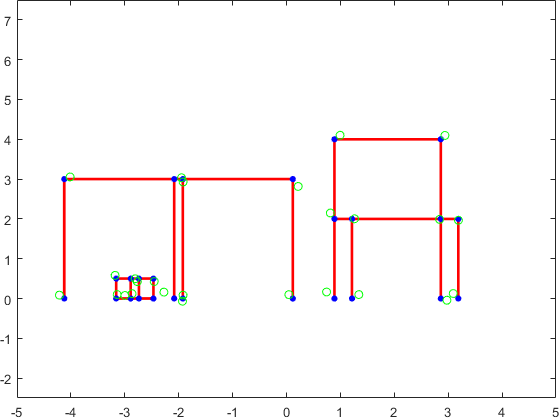}
	\end{center}
	\caption{Examples of synthetic 3D scences and images. First row: 3 randomly placed objects and the camera location. Second row: Orthographic projections (blue poines and read lines) corresponding to the above scenes, and image points (green circles) perturbed by Gaussian noise.}
	\label{fig:synt_data}
\end{figure*}

In this section we further discuss the matrix model given in Section~\ref{sec:matrix_model} and give a concrete example with orthogonal projection of a scene consisting of three rigidly moving objects.
Figure~\ref{fig:synt_data} shows some scene examples and the images resulting from the projection.

Suppose that $S$ is a $3 \times p$ matrix containing column vectors representing the coordinates of the $p$ points of one of the scene objects. We rigidly move the object in the $xy$-coordinate plane by computing
$W = RS+t\one$, where 
\begin{equation}
	R = \left(\begin{matrix}
		a & -b & 0 \\
		b & a & 0 \\
		0 & 0 & 1
	\end{matrix}\right),
	\quad
	t = \left(\begin{matrix}
		t_1 \\
		t_2 \\
		0
	\end{matrix}\right)
\end{equation}
and $1$ is a $1\times p$ vector of all ones. Note that $W$ is of the same size as $S$.
The row vector $y_i$ which represents scene will contain the column-stacked rows $w_1, w_2, w_3$ of $W$. 
If $s_1$,$s_2$ and $s_3$ are the rows of $S$ we define the block matrix
\begin{equation}
	\bar{S} = \begin{bmatrix}
		s_1 & \zero & \zero \\
		s_2 & \zero & \zero \\
		\zero & s_1 & \zero \\
		\zero & s_2 & \zero \\
		\zero & \zero & s_3 \\
		\one & \zero & \zero \\
		\zero & \one & \zero 
	\end{bmatrix},
\end{equation}
where $\zero$ is a $1 \times p$ row vector of all zeros.
We can now write 
\begin{equation}
	\begin{bmatrix}
		w_1 & w_2 & w_3
	\end{bmatrix}
	= v \bar{S},
\end{equation}
where $v$ contains the transformation parameters
\begin{equation}
	v = \left(\begin{matrix}
		a & -b & b  & a & 1 & t_1 & t_2
	\end{matrix}\right).
	\label{eq:paramvector}
\end{equation}
Now for multiple transformations we collect their parameters in row vectors $v_j$ as in \eqref{eq:paramvector}, and column-stack them into a matrix $V$. The object pose in a particular scene is then represented by the corresponding row of the matrix $V\bar{S}$. It is clear that the matrix $\bar{S}$ has rank $7$ unless the $s_1$,$s_2$,$\one$ are linearly dependent. On the other hand it is easy to see that the row space of $V$ is spanned by a 5-dimensional basis. Therefore the rank of $V\bar{S}$ is $5$.

For each of the three objects in the scene we now create structure- $\bar{S}_1,\bar{S}_2,\bar{S}_3$ and parameter-matrices $V_1, V_2, V_3$.
The block matrix $Y=\begin{bmatrix}
V_1\bar{S}_1 & V_2\bar{S}_2 & V_3\bar{S}_3
\end{bmatrix}$ which contains the coordinates from all three objects can then be written
\begin{equation}
	Y = \begin{bmatrix}
		V_1 & V_2 & V_3
	\end{bmatrix}
	\begin{bmatrix}
		\bar{S}_1 & 0 & 0 \\
		0 & \bar{S}_2 & 0 \\
		0 & 0 & \bar{S}_3
	\end{bmatrix}.
\end{equation}
The matrix containing the structure is clearly of full rank as long as $\bar{S}_1,\bar{S}_2$ and $\bar{S}_3$ all have full rank.
However $V$ contains three columns with all ones and therefore the rank is reduced by when concatenating $V_1$,$V_2$ and $V_3$. This gives $\text{rank}(Y)=3\cdot5-2=13$. 

The image matrix $X$ is generated by applying a matrix $\Pi$ which simply discards columns of $Y$ that correspond to depth (all $y$-coordinates in our application).

For each object this corresponds to applying a matrix $\Pi$ such that
\begin{equation}
	\begin{bmatrix}
		s_1 & \zero & \zero \\
		s_2 & \zero & \zero \\
		\zero & s_1 & \zero \\
		\zero & s_2 & \zero \\
		\zero & \zero & s_3 \\
		\one & \zero & \zero \\
		\zero & \one & \zero 
	\end{bmatrix} \Pi = 
	\begin{bmatrix}
		s_1 &  \zero \\
		s_2 &  \zero \\
		\zero & \zero \\
		\zero & \zero \\
		\zero & s_3 \\
		\one  & \zero \\
		\zero  & \zero 
	\end{bmatrix}.  
\end{equation}
We can therefore write 
\begin{equation}
	X = \begin{bmatrix}
		V_1\bar{S}_1\Pi & V_2\bar{S}_2 \Pi & V_3\bar{S}_3 \Pi 
	\end{bmatrix}= Y \bar{\Pi},
\end{equation}
where
\begin{equation}
	\bar{\Pi} = \begin{bmatrix}
		\Pi & 0 & 0 \\
		0 & \Pi & 0 \\
		0 & 0 & \Pi
	\end{bmatrix}.
\end{equation}
Note that the columns of $V_i$ that corresponds to $t_2$ completely disappear from the product $V_i \bar{S}_i \Pi$ these can therefore be assumed to be $0$. This explains why an objects distance to the camera cannot be recovered from $X$. With this element set to zero the rank of $V_i$ is now $4$. The concatenation 
$\begin{bmatrix}
V_1 & V_2 & V_3
\end{bmatrix}$
still contains three columns with all ones. We therefore get $\text{rank}(V) = 3 \cdot 4-2 = 10$, which similar to above can easily be seen to be the rank of $X$ as well.

The final step of generating the images is to add noise to the image coordinates. This results in a full rank matrix
\begin{equation}
	\tilde{X} = X + N,
\end{equation} 
where $N$ is a matrix that contains Gaussian elements.
In the following sections we will consider the effects of trying to learn this model with and without auxiliary variables.

\subsection{Learning Without Z-variabels}
\begin{figure}[htb]
	\begin{center}
		\includegraphics*[width=70mm]{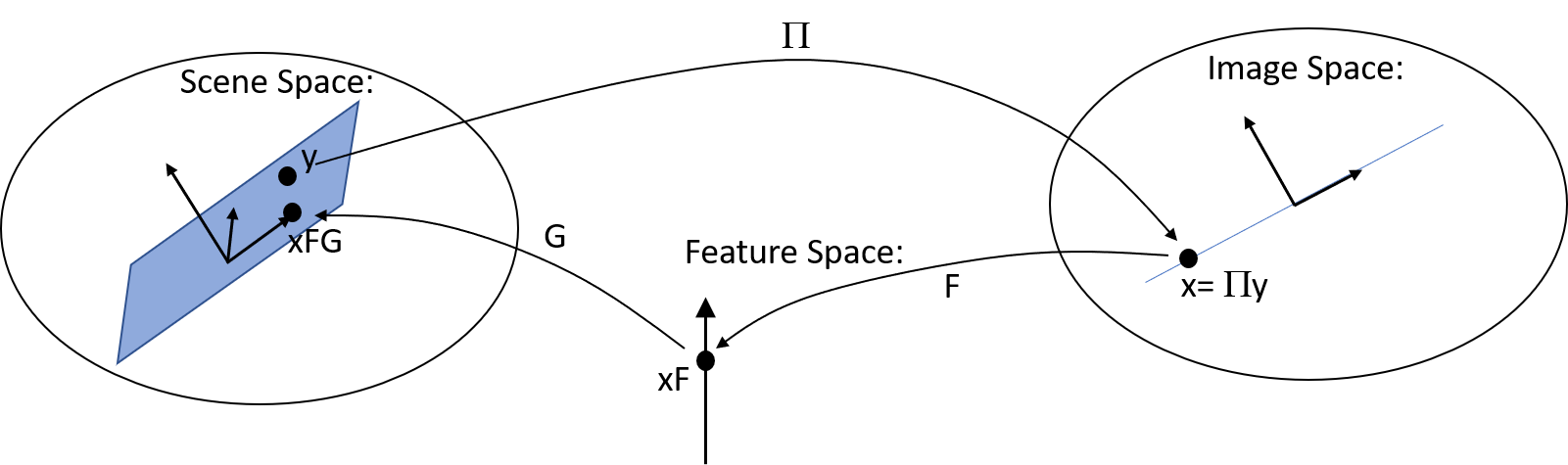}
	\end{center}
	\caption{Illustration of the mappings involved when learning without a parameterization of the knowledge gap.}
	\label{fig:linspaces_no_z}
	\begin{center}
		\includegraphics*[width=70mm]{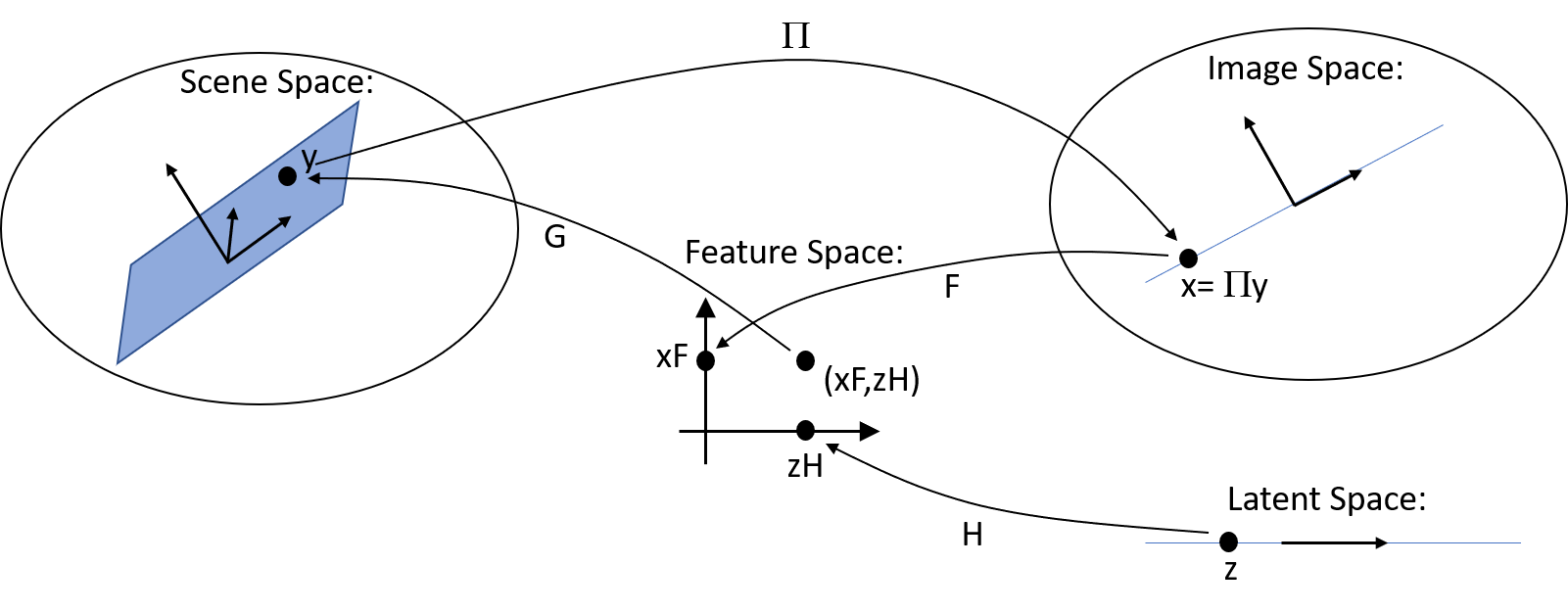}
	\end{center}
	\caption{Illustration of the mappings involved when learning with a parameterization of the knowledge gap.}
	\label{fig:linspaces_z}
	
\end{figure}

Recall the network construction from \eqref{eq:linmodel}. In this section we consider the case where $Z=0$ and apply it to the orthographic projection problem described above. 
Figure~\ref{fig:linspaces_no_z} illustrates the mappings and spaces involved in the problem.
Each scene is represented by a row vector with $78$ elements, corresponding to the coordinates of all the scene points. The values of these $78$ variable are however not arbitrary but belongs to 13-dimensional linear subspace that contains all the scenes we can expect encounter.
Thus our scene space is a subspace of $\mathbb{R}^{78}$.
Similarly the images are represented by row vectors of size $52$ and the feasible images belong to the $10$-dimensional image space. The forward projection $\bar{\Pi}: \mathbb{R}^{78} \mapsto \mathbb{R}^{52}$ maps scenes to images. From elementary linear algebra it is clear that it cannot be bijective due to the sizes of the subspaces. (As we saw in the previous section the missing information required to invert the mapping is one  absolute depth for each object.) Applying $F: \mathbb{R}^{52} \rightarrow \mathbb{R}^{10}$ to an image vector $x$ gives a representation $xF$ in the feature space. To get back to the scene space we apply $G_x: \mathbb{R}^{10} \rightarrow \mathbb{R}^{72}$ to $xF$ giving the point $x F G_x$ in the scene space.
Note that since the feature space has dimension $10$ there is no way we can recover the whole scene space, but we have to settle for a $10$ dimensional subspace. 

\begin{figure*}
	\begin{center}
		\def\w{30mm}
		\includegraphics[width=\w]{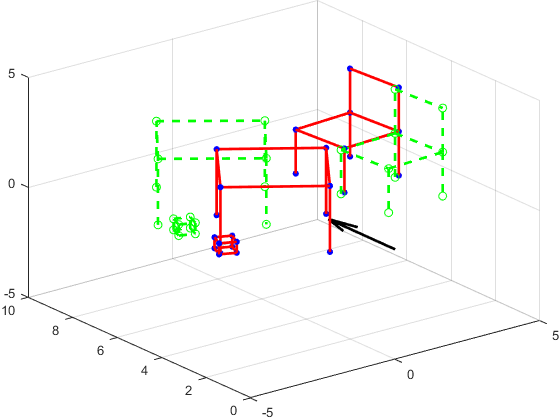}			
		\includegraphics[width=\w]{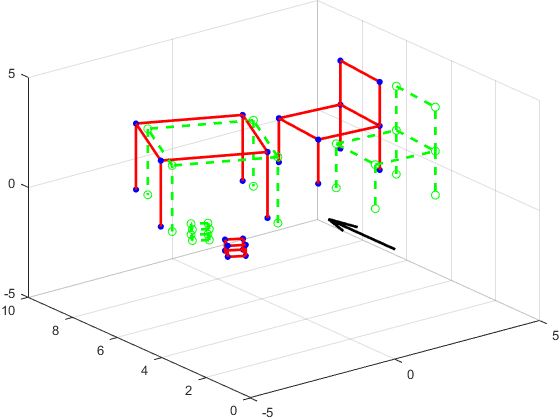}			
		\includegraphics[width=\w]{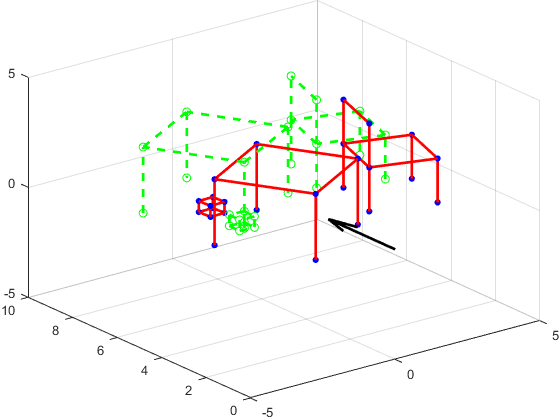}			
		\includegraphics[width=\w]{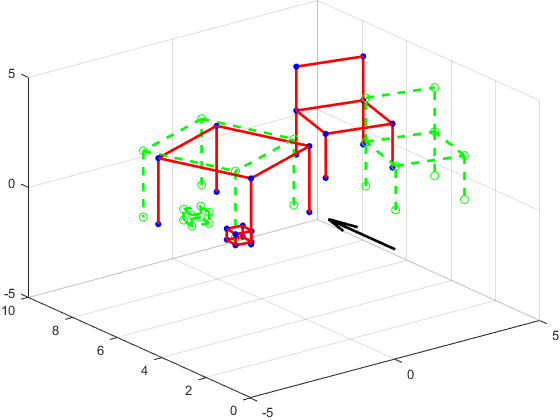}			
	\end{center}
	\caption{Depth predictions on test data without $z$ variables. Green dashed lines is the prediction, red lines and blue points is the true scene.}
	\label{fig:predictions_no_z}
	\begin{center}
		\def\w{30mm}
		\includegraphics[width=\w]{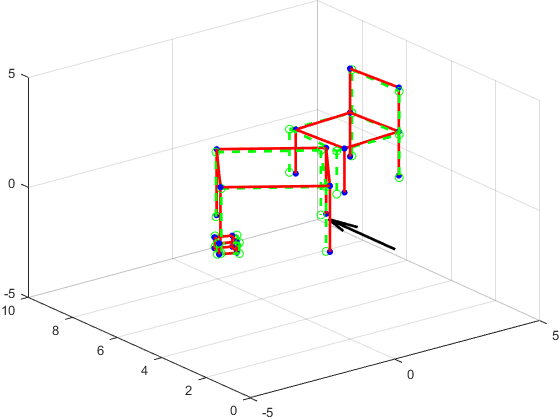}			
		\includegraphics[width=\w]{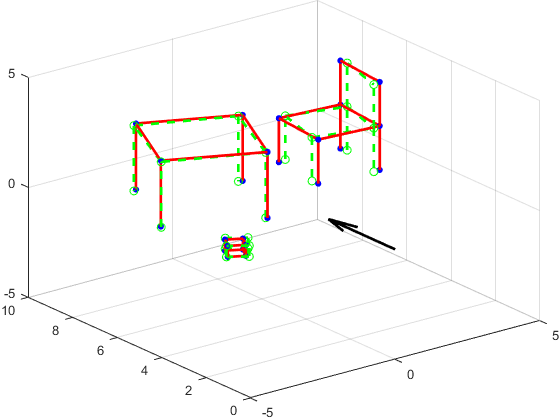}			
		\includegraphics[width=\w]{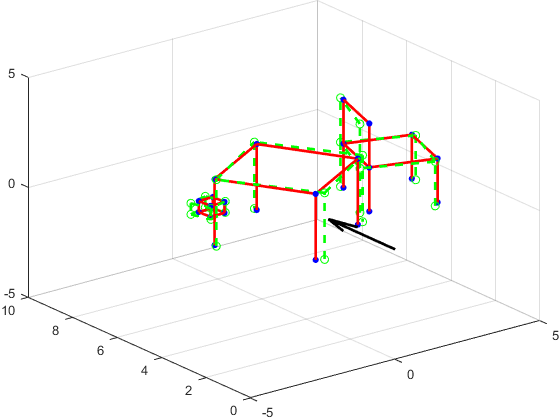}			
		\includegraphics[width=\w]{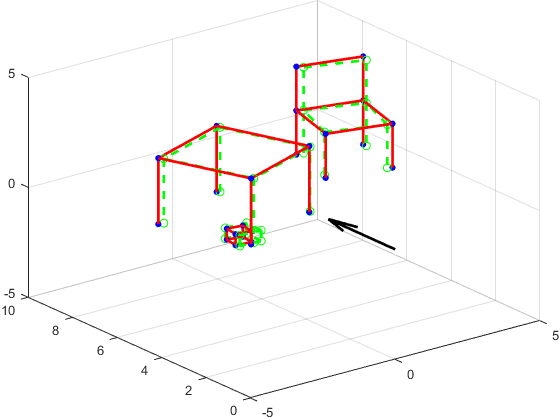}			
	\end{center}
	\caption{Depth predictions on test data with optimization over $z$ variables. 
		Green dashed lines is the prediction, red lines and blue points is the true scene.}
	\label{fig:predictions_z}
\end{figure*}

To train the model we generated $50$ ground truth scenes $Y_\text{train}$ and corresponding images $X_\text{train}$ as described in the previous section. Since the training data has noise we use the loss function
\begin{eqnarray}
	\hspace{-5mm}& \min_{X,F,G_x} &  \|XFG_x - Y_\text{train}\|_F^2 + \|X-X_{\text{train}}\|_F^2 \\
	\hspace{-5mm}& \text{such that} &  \text{rank}(X) = 10.
\end{eqnarray}
The additional matrix $X$ contains a clean estimate of the image subspace spanned by the rows of $X_\text{test}$. Note that the first term $\|XFG_x - Y_\text{train}\|_F^2$ only depends on the rank of $X$ and not which particular subspace it spans. Therefore the optimal $X$ will be given by
\begin{equation}
	[X]_{10} = \argmin_{\text{rank}(X) = 10} \|X-X_{\text{train}}\|_F^2,
\end{equation}
where $[X]_{10}$ is obtained by thresh holding the singular values of $X_\text{train}$.
The above loss is therefore equivalent to
\begin{equation}
	\| [X]_{10} F G_x -Y_\text{train}\|_F^2.
\end{equation}
In Figure~\ref{fig:predictions_no_z} we compare a few of the scenes predicted from the test data to the true scene. While the rotational pose of the objects is well predicted their depths are as expected not correct.

Since $X_\text{train}$ is corrupted by noise it is in fact of full rank. Therefore it is possible to try the naive approach of making the loss
\begin{equation}
	\|X_\text{train} F G_x - Y_\text{train}\|_F^2
\end{equation}
completely vanish by increasing the size of the feature space to 13.
This will however result in overfitting since the 3-extra dimensions will depend purely on noise.
This is confirmed by the results in presented in Table~\ref{tab:overfitting}.
\begin{table}[htb]
	\begin{center}
		\begin{tabular}{c|cc}
			Feature Space: & Training data: & Test data: \\
			\hline
			10D (without Z) & 44.4252 & 31.2420 \\
			13D (without Z)& 0 & 130.4383 \\
			\hline
			(10+3)D (with Z) & 7.3618 & 2.8746
		\end{tabular}
	\end{center}
	\caption{Sum-of-squares distances between predictions to true scene points, for both training and test data.}
	\label{tab:overfitting}
\end{table}

\subsection{Learning With Z-variables}
We now add the shape parameterization to the problem. 
Rather than predicting a single point in the scene space from an image we are now interested in finding the whole subspace of scenes, parametrized by $z$, that project to the viewed image. 
Figure~\ref{fig:linspaces_z} illustrates our approach. In contrast to the previous section we add the latent variabels $z$ to the feature space increasing its dimensionality to $13$ making it possible to pinpoint a specify unique point in the scene space that projects to the viewed image. 

We consider the objective
\begin{equation}
	\|XFG_x + ZG_z - Y_\text{train}\|_F^2 + \|X-X_{\text{train}}\|_F^2,
\end{equation}
where $\text{rank}(X) = 10$. Similar to above the first term only depends on the rank of $X$ and not which particular subspace it spans. Therefore optimization over $X$ reduces the problem to
\begin{equation}
	\|[X]_{10} F G_x + ZG_z - Y_\text{train}\|_F^2.
\end{equation}
We now divide $Y_\text{train}$ into two components; $\mathcal{P}Y_\text{train}$ which is the projection of $Y_\text{train}$ onto the columnspace of $[X]_{10}$, and $\mathcal{P_\perp} Y_\text{train}$ which is perpendicular to it. Letting $Z$ have columns that are perpendicular to $[X]_{10}$ gives 
\begin{equation}
	\|[X]_{10} F G_x + \mathcal{P}Y_\text{train}\|^2_F + \|ZG_z - \mathcal{P}_\perp Y_\text{train}\|_F^2.
\end{equation}
Since $[X]_{10}$ spans the columns of $\mathcal{P} Y_\text{train}$ it is clear that the first term can be made to vanish by selecting $FG_x = [X]_{10}^\dagger \mathcal{P}Y_\text{train}$.
The second term is minimized when $Z G_z$ is the best rank $3$ approximation of $\mathcal{P_\perp}Y_\text{train}$, which can be computed from the SVD of $\mathcal{P_\perp}Y_\text{train}$.

Figure~\ref{fig:predictions_z} shows the result of applying the above training approach to the orthographic projection problem. Here we plot the best fit to the true 3D scene that was used to generate the test image.
Note that given the image we get different scene predictions when varying $z$.
Here we have optimized over the $z$ to find the scene that is as close to true scene as possible.
In the last row of Table~\ref{tab:overfitting} we also display the sum-of-squares distance between the predictions and the scenes. It is clear that the introduction of the z-variables gives a shape model that captures all the variations in the training data even if the images themselves are not enough to predict them.
In a situation where the true scene shape is unknown the free parameters could for example be fixed using a second camera in a stereo setup. 

\section{Selecting Co-Visible Images \label{sec:covisible_selection}}

To be able to train the network or estimate depth maps using self supervision, we need to have sets of visually overlapping images, called co-visible images. These images must have certain characteristics to be informative, such as sufficient overlap, parallax and low redundancy. Redundancy here means images that are captured from approximately the same view points.

When using images from structure from motion solutions, as in~\cite{enqvist-etal-omnivs-2011}, we assume that all images have been captured with sufficient relative parallax and low redundancy. We therefore simply group images by selecting a reference image and then selecting $N-1$ other images that observe the most 3d points in common with the reference image. 

The Scannet dataset~\cite{dai2017scannet} comprises of sequences of images accompanied with ground truth depth maps and camera poses. This means that we do not have direct knowledge of visual overlap, and since they are captured close in time, the most visually overlapping view will be very redundant, with very low parallax.

To select co-visible images from this dataset, we first find which images have visual overlap. This is done by creating a voxel map, were a camera id is added to a voxel if its un-projected depth map intersects with the voxel. This is done in one pass for all cameras. Next a second pass is done, where each camera fetches the set of cameras that intersected with the same voxels it does. This set becomes the set of overlapping cameras. This allows us to find the overlapping cameras in linear time, rather than cubic time.

In the next step we create the co-visible set for each camera. The set initially contains only the camera itself. We then iteratively add new cameras to the set from the set of overlapping cameras. In each iteration we select the camera that has the largest parallax to all cameras already in the set and with sufficiently high overlap to all of the cameras.

\subsection{Evaluation of Training Approach}
While self supervision gives access to more training data, the quality of that data may not be as high as ground truth depth maps.
In this section we therefore compare our two supervision approaches to ensure that we can achieve similar results. 

Using the same datasets as in Section~\ref{sec:supervisedtraing}, we extract sets of five co-visible images as described in~\ref{sec:covisible_selection}. The amount of sets is then sub-sampled to $30000$ by drawing them at random, yielding in total $150000$ images. We do this to have the same amount of training data in both cases.

Table~\ref{table:traing} shows that with this data, supervised and self-supervised training yield comparable results, with the supervised training giving slightly better metrics. Still, we are able to achieve very good performance with a the self supervised approach which is scalable and can benefit from the large amount of available structure from motion datasets.

\begin{table}[!htbp]
\begin{center}
\resizebox{0.48\textwidth}{!}{\begin{tabular}[h]{|l|l|c c|c c|}
\hline
 & & \multicolumn{2}{|c|}{lower is better} & \multicolumn{2}{|c|}{higher is better}\\
Dataset & Method
&absrel  &log RMSE$^{sc.}_{inv.}$& $\delta<1.1$ & $\delta<1.25$ \\ 
\hline\hline
\textbf{scannet/scene0707\_00}&Supervised & 0.0805 & 0.0071  &73.50\%& 93.44\% \\ 
&Self-supervised & 0.0957 &  0.0095 & 66.65\% & 90.56\%\\ 
\hline
\textbf{scannet/scene0715\_00}&Supervised & 0.0582&0.0045& 83.40\% & 96.67\% \\
&Self-supervised & 0.0735  & 0.0760 & 76.25\% & 93.54\% \\
\hline
\textbf{scannet/scene0799\_00}&Supervised & 0.0658& 0.0056& 79.43\% & 95.00\%\\ 
&Self-supervised & 0.0791& 0.0084& 73.48\%& 91.84\% \\
\hline
\textbf{TUM/seq1}&Supervised &  0.1261 & 0.0164& 72.43\% & 83.56\% \\
&Self-supervised& 0.2528 & 0.0437 &62.38\% & 71.33\% \\
\hline
\textbf{TUM/seq2}&Supervised & 0.0409& 0.0063& 90.56\% & 95.57\% \\ 
&Self-supervised& 0.0575& 0.0102& 88.23\% & 92.22\%\\
\hline
\textbf{TUM/seq3} &Supervised & 0.0910 & 0.0187& 74.93\% & 88.25\%\\ 
&Self-supervised & 0.1086& 0.0264& 71.42\% & 85.69\%\\
\hline
\end{tabular}} 
\caption{Evaluation of training approach, comparing networks that have been trained supervised and self-supervised.
}
\label{table:traing}
\end{center}
\end{table}

\section{Description of training data}
Here we present the training data used to train the network for the experimental comparison in Section~\ref{sec:comparison_to_others}. We used two sources for our training dataset. Firstly, the Scannet training datasets (v2)~\cite{dai2017scannet} where every dataset with even number was used, only considering the sequence ending with $\_00$ therein. Secondly, a subset of the structure from motion datasets presented  in~\cite{enqvist-etal-omnivs-2011}, with the following names;
\begin{figure*}[ht!]
    \centering
     \begin{subfigure}[b]{0.31\textwidth} 
 \centering
\resizebox{\textwidth}{!}{
  \includegraphics[width = 1.3cm]{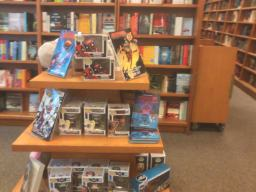}
     \includegraphics[width = 1.3cm]{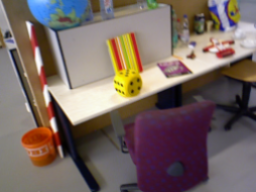}
      \includegraphics[width = 1.3cm]{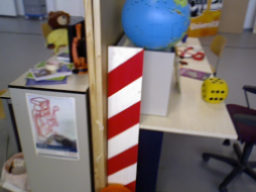}}
\\
 \vspace{-0.5cm}
     \resizebox{\textwidth}{!}{ 
       \resizebox{\textwidth}{!}{\hspace{0.1cm}\tiny{\alert{\textbf{scannet/scene0799\_00}}}\hspace{0.5cm}
       \alert{\textbf{TUM/seq3}}\hspace{1.3cm}
       \alert{\textbf{TUM/seq3}}\hspace{1.2cm}}}
       \\
       \vspace{0.05cm}
      \resizebox{\textwidth}{!}{  \includegraphics[width = 1.3cm]{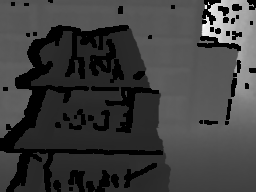}
        \includegraphics[width = 1.3cm]{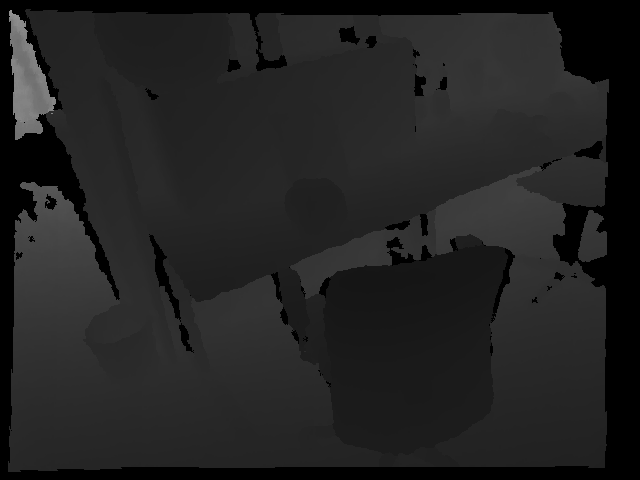}
      \includegraphics[width = 1.3cm]{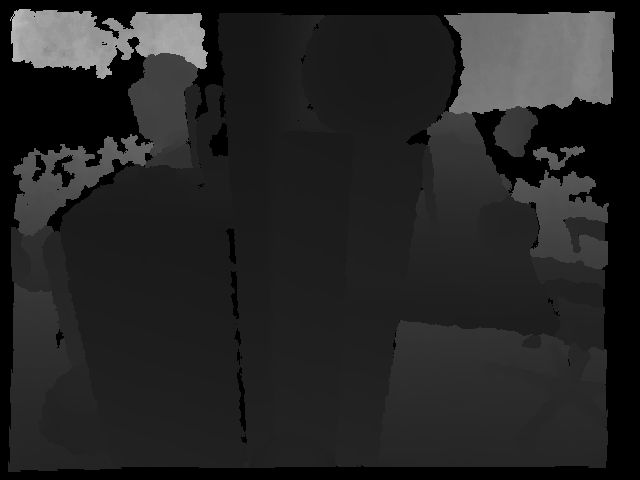}}
      \\ 
       \resizebox{\textwidth}{!}{ \includegraphics[width = 1.3cm]{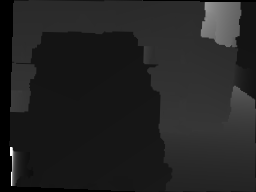}
        \includegraphics[width = 1.3cm]{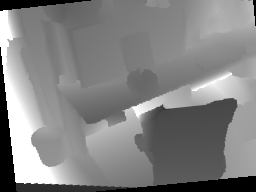}
      \includegraphics[width = 1.3cm]{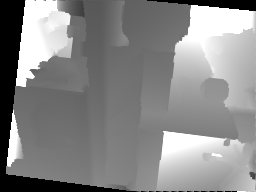}}
\\ 
\resizebox{\textwidth}{!}{\hspace{0.1cm}\includegraphics[width = 1.3cm]{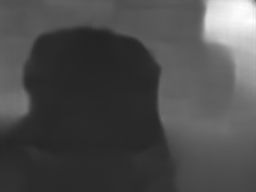}
       \includegraphics[width = 1.3cm]{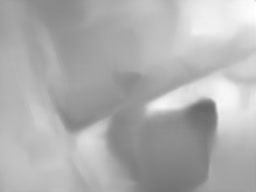}
      \includegraphics[width = 1.3cm]{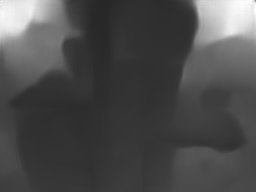}\hspace{-0.1cm} } \vspace{-2.3cm}
     \end{subfigure}
     \hspace{0.3cm}
    \begin{subfigure}{0.66\textwidth}
    \vspace{-0.3cm}
        \resizebox{\textwidth}{!}{
     \includegraphics[width = 18mm, trim = 100 200 100 100]{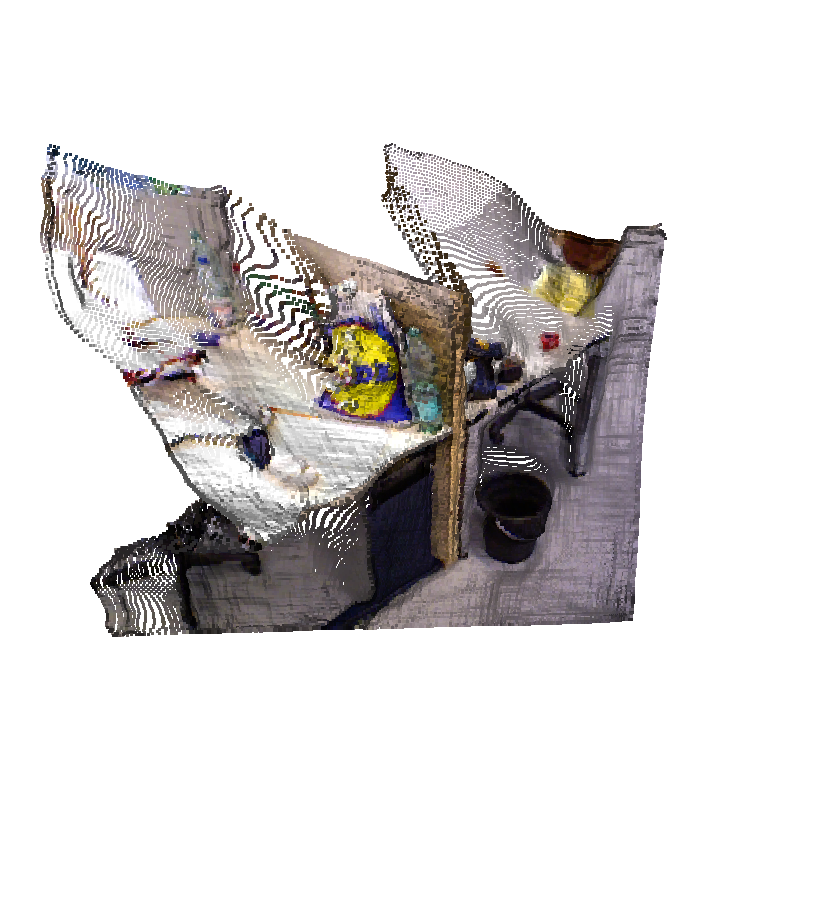}
 \includegraphics[width = 18mm, trim = 100 150 100 80]{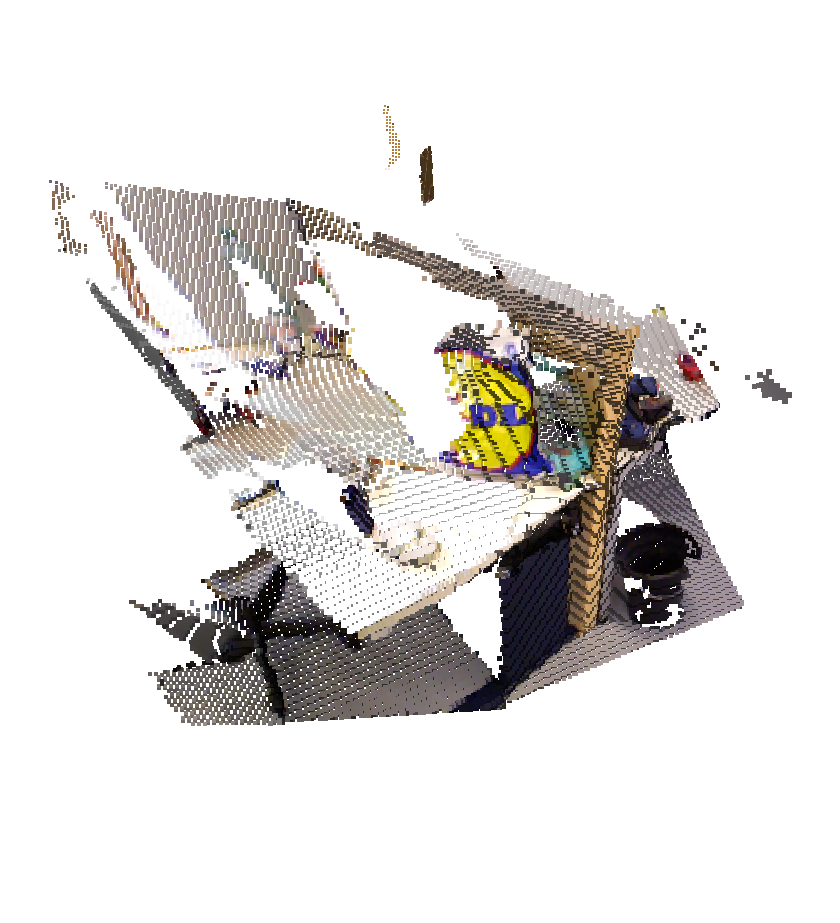}\hspace{0.3cm}
 \includegraphics[width = 18mm, trim = 50 150 50 50]{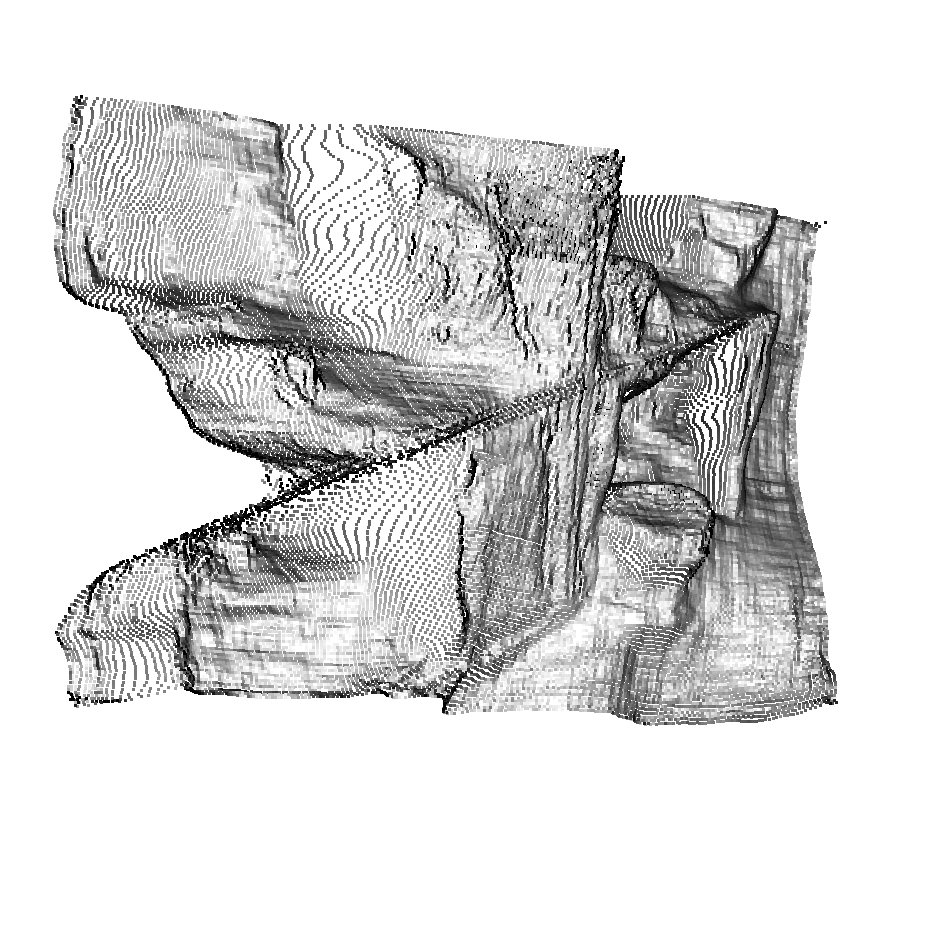}
 \includegraphics[width = 18mm, trim = 50 150 50 50]{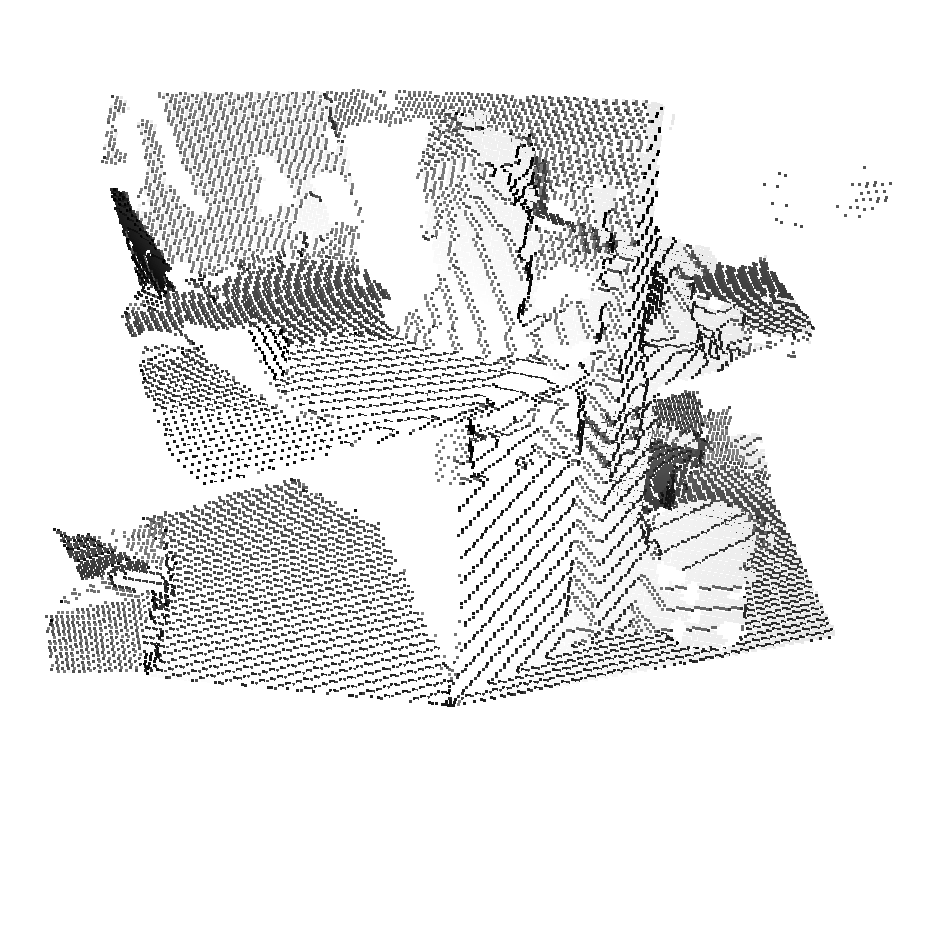}
}\\
   \vspace{0.5cm}
   \resizebox{\textwidth}{!}{\hspace{2cm}\footnotesize{\textbf{Ours}}\hspace{2cm}\footnotesize{\textbf{Stereo}}
   \hspace{2cm}\footnotesize{\textbf{Ours}} \hspace{2cm}\footnotesize{\textbf{Stereo}}}
    \vspace{0.2cm}\\
   \resizebox{\textwidth}{!}{
    \includegraphics[width = 18mm, trim = 40 230 40 300]{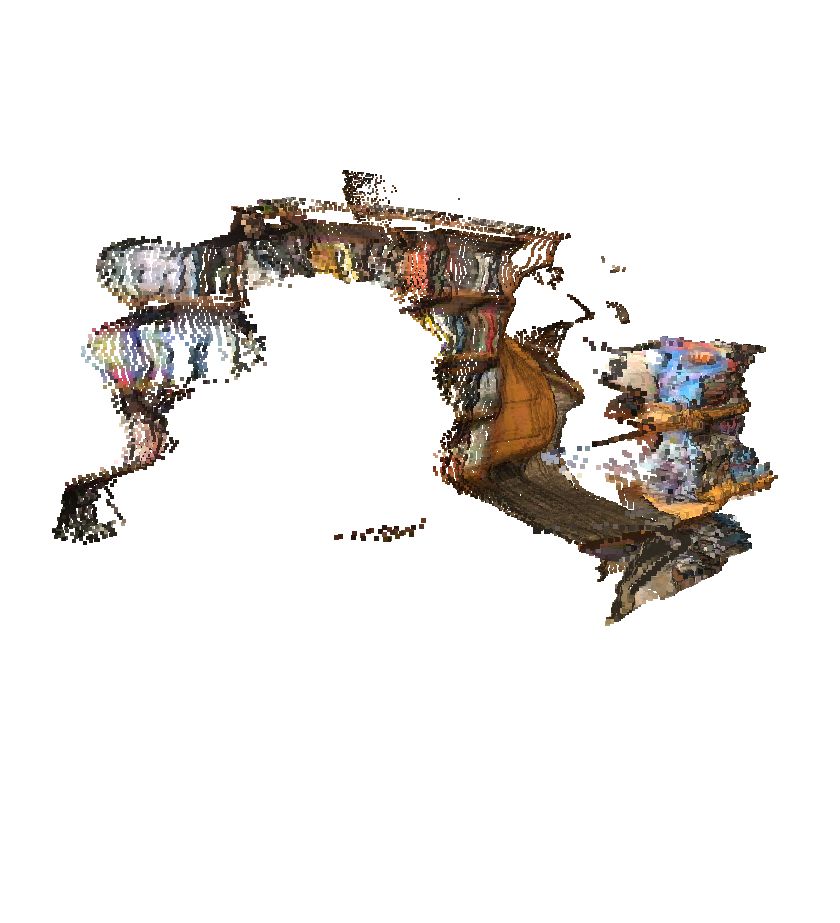}
  \includegraphics[width = 19mm, trim = 40 200 40 300]{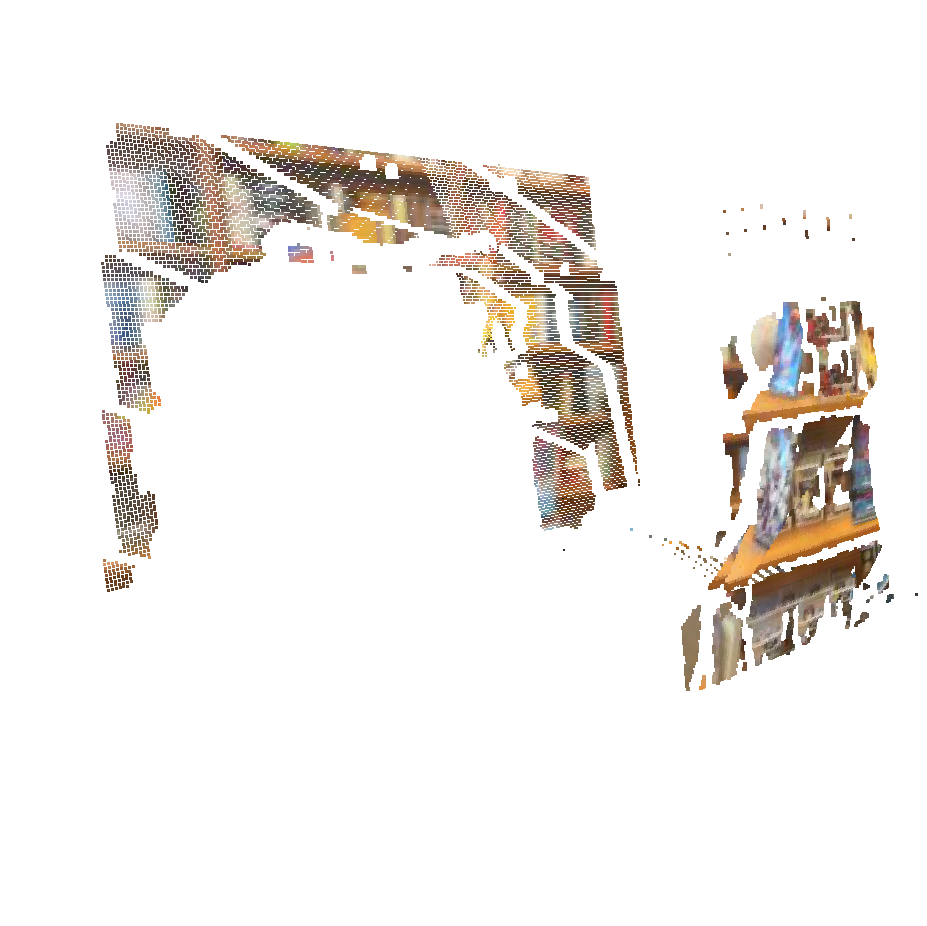}
   \includegraphics[width = 19mm, trim = 40 230 40 300]{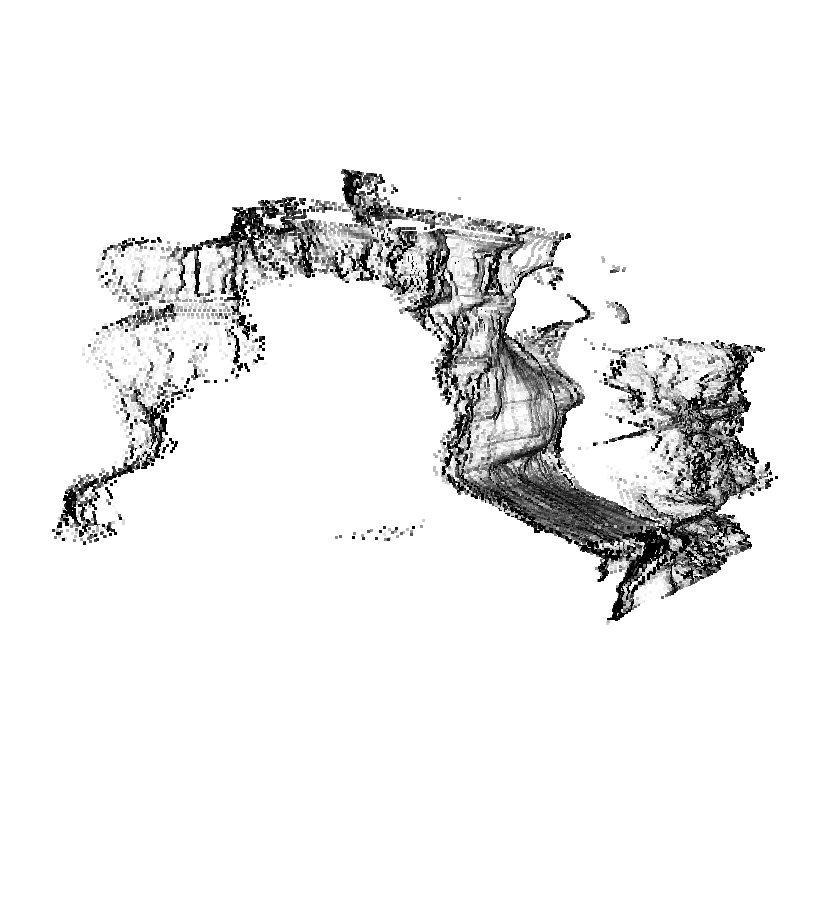}
   \includegraphics[width = 18mm, trim = 40 200 40 200]{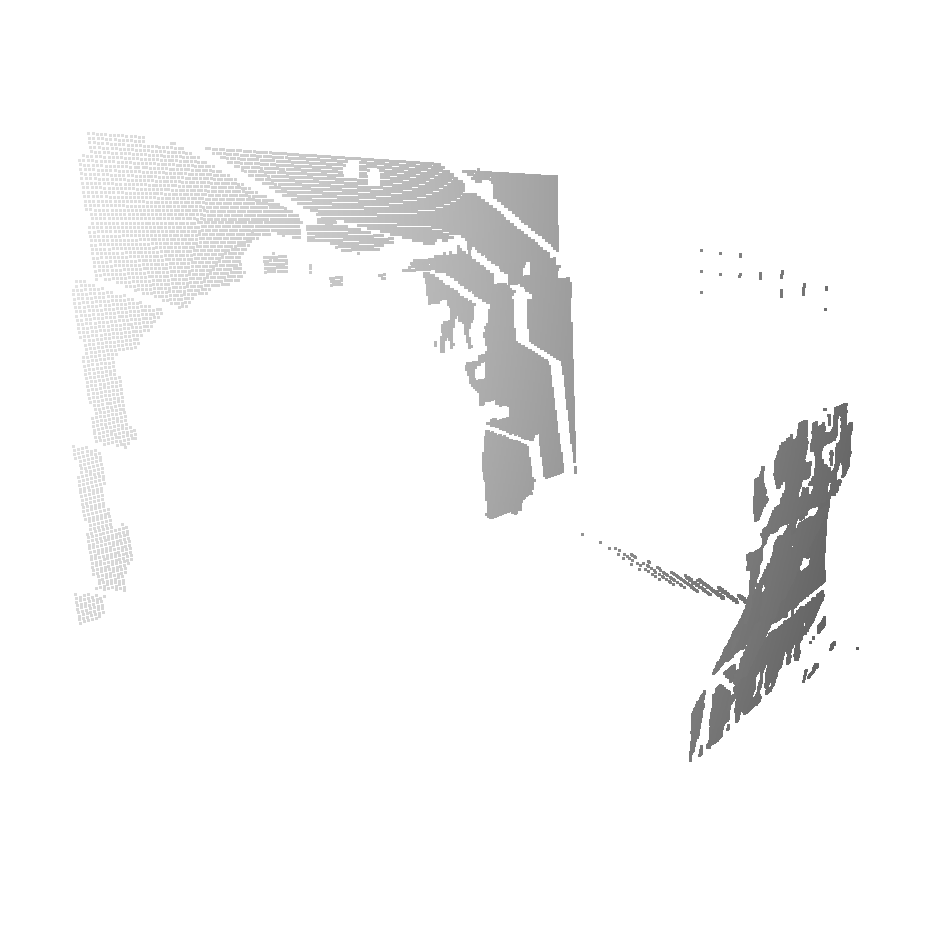}
} \vspace{-1cm}
\\
\\
\\
   \resizebox{\textwidth}{!}{\hspace{2cm}\footnotesize{\textbf{Ours}}\hspace{2cm}\footnotesize{\textbf{Stereo}}
   \hspace{2cm}\footnotesize{\textbf{Ours}} \hspace{2cm}\footnotesize{\textbf{Stereo}}}
  \vspace{1.0cm}
        \label{fig:quantitative_othermethods3d}
            \end{subfigure}
            \vspace{-1.5cm}
   \caption{\textbf{Stereo comparison.} Comparison of depth maps produced by stereo method and ours. The left images correspond to input image , ground truth, stereo depth and Photometric (ours) (from upper to lower). The right images shows reconstruction two of the three depth estimations for ours and Stereo method.}
   \label{fig:stereocomp}
\end{figure*}
\begin{figure*}
   \resizebox{\textwidth}{!}{
   \includegraphics[]{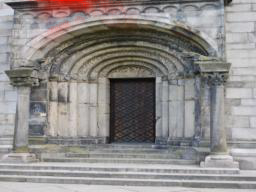}
   \includegraphics[]{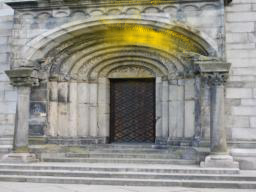}
   \includegraphics[]{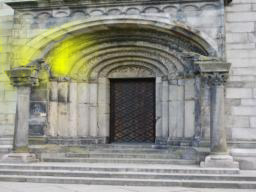}
 \includegraphics[]{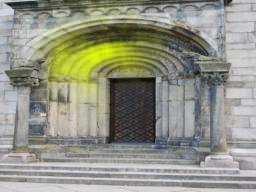}
 \includegraphics[]{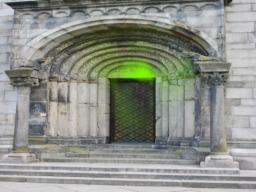}
 \includegraphics[]{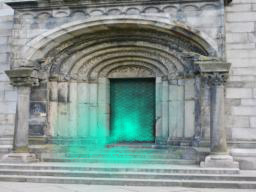}
 \includegraphics[]{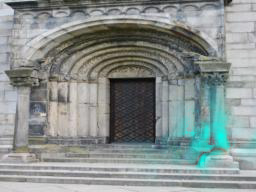}
 \includegraphics[]{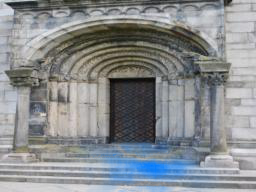}
   }\\
   \resizebox{\textwidth}{!}{
  \includegraphics[]{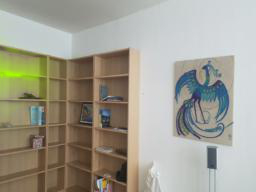}
  \includegraphics[]{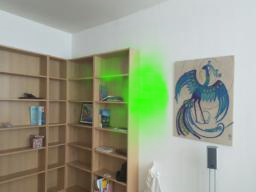}
    \includegraphics[]{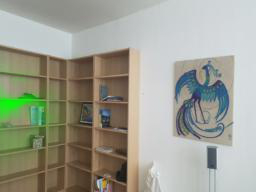}
      \includegraphics[]{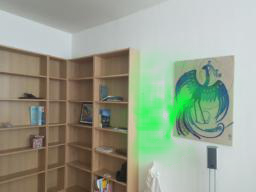}
        \includegraphics[]{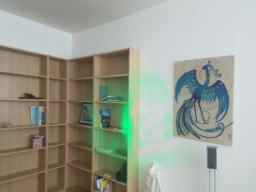}
 \includegraphics[]{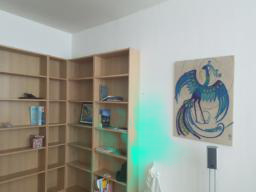}
  \includegraphics[]{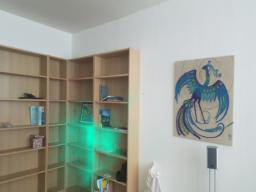}
 \includegraphics[]{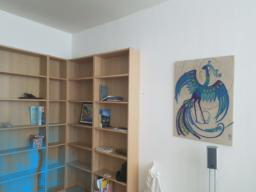}
   }
   \\
   \resizebox{\textwidth}{!}{
  \includegraphics[angle = 90]{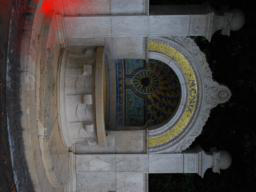}
  \includegraphics[angle = 90]{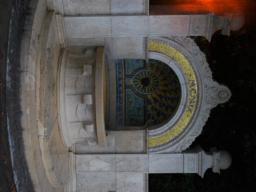}
    \includegraphics[angle = 90]{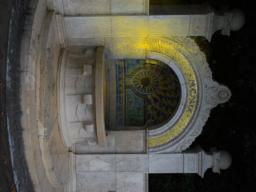}
      \includegraphics[angle = 90]{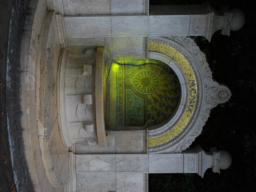}
        \includegraphics[angle = 90]{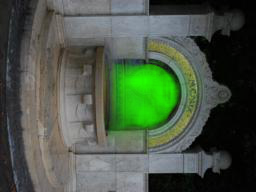}
          \includegraphics[angle = 90]{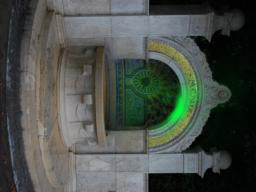}
            \includegraphics[angle = 90]{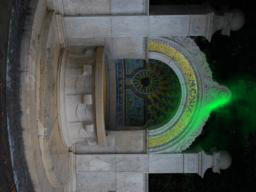}
              \includegraphics[angle = 90]{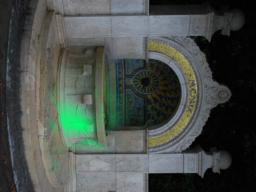}
                \includegraphics[angle = 90]{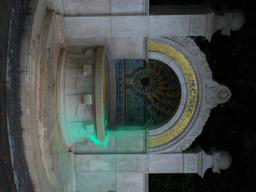}
                  \includegraphics[angle = 90]{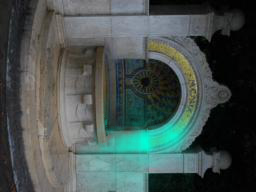}
   }
\caption{\textbf{Local influence of $z$-variable}. Each figure shows the local effect of a small shift in one dimension of the $z$-variable. The change is coded in color.}
\label{fig:latentvar}
\end{figure*}
\begin{itemize}
	\setlength\itemsep{0.01em}
	\item Modern Museum Ceiling, Barcelona
	\item Ecole Superior De Guerre
	\item Doge Palace, Venice
	\item Door, Lund
	\item Fort Channing gate, Singapore
	\item Kings College Interior, Cambridge
	\item Staircase 1, Doge Palace, Venice
	\item Staircase 2, Doge Palace, Venice
	\item Council Chamber, Doge Palace, Venice
	\item Doge Palace, Venice
	\item Cathedral ceiling, Barcelona
	\item Eglise du dome
	\item Drinking Fountain, Z\"urich
	\item Thian Hook Keng temple, Singapore
\end{itemize}

\section{Stereo comparison}
In this section we present depth maps of the two models compared in Section 3.1 \textit{Ablation studies, Stereo comparison}. In Figure \ref{fig:stereocomp} we show examples of how our model captures the geometrical shapes of the objects more correctly than the stereo method. The stereo method from \cite{olsson-etal-cvpr-2013} tend to put the objects into planar shapes.

\section{Effects of Latent Variables}

In Figure~\ref{fig:latentvar} more examples of the impact of the individual $z$-components are displayed. As stated before, the $z$-components have a local impact on the depth map, where the shape and size  depend on the image.

{\small
}

\end{document}